\documentclass{article}

\usepackage{microtype}
\usepackage{graphicx}
\usepackage{subfigure}
\usepackage{booktabs} 
\usepackage{amssymb}
\usepackage{multirow}

\usepackage{hyperref}

\usepackage{color}
\usepackage{xspace}
 
\usepackage[accepted]{icml2025}

\usepackage{amsmath}
\usepackage{amssymb}
\usepackage{mathtools}
\usepackage{amsthm}

\usepackage{float}

\usepackage[capitalize,noabbrev]{cleveref}

\theoremstyle{plain}

\theoremstyle{definition}

\theoremstyle{remark}

\usepackage[textsize=tiny]{todonotes}

\icmltitlerunning{LocalEscaper: A Weakly-supervised Framework with Regional Reconstruction for Scalable Neural TSP Solvers}

\begin{document}

\twocolumn[
\icmltitle{LocalEscaper: A Weakly-supervised Framework with Regional Reconstruction for Scalable Neural TSP Solvers}




\begin{icmlauthorlist}
\icmlauthor{Junrui Wen}{hust}
\icmlauthor{Yifei Li}{hust}
\icmlauthor{Bart Selman}{Cornell} 
\icmlauthor{Kun He}{hust}
\end{icmlauthorlist}

\icmlaffiliation{hust}{School of Computer Science, Huazhong University of Science and Technology, China}
\icmlaffiliation{Cornell}{Department of Computer Science, Cornell University, United States}

\icmlcorrespondingauthor{Kun He}{brooklet60@hust.edu.cn}

\icmlkeywords{combinatorial optimization, neural solver, weakly-supervised learning, TSP}

\vskip 0.3in
]



\footnotetext[1]{School of Computer Science, Huazhong University of Science and Technology, China. ${}^2$Department of Computer Science, Cornell University, United States. Correspondence to: Kun He $<$brooklet60@hust.edu.cn$>$.\\\\Preprint.}


\begin{abstract}
Neural solvers have shown significant potential in solving the Traveling Salesman Problem (TSP), yet current approaches face significant challenges. Supervised learning (SL)-based solvers require large amounts of high-quality labeled data, while reinforcement learning (RL)-based solvers, though less dependent on such data, often suffer from inefficiencies. 
To address these limitations, we propose LocalEscaper, a novel weakly-supervised learning framework for large-scale TSP. 
LocalEscaper effectively combines the advantages of both SL and RL, enabling effective training on datasets with low-quality labels. To further enhance solution quality, we introduce a \textbf{regional reconstruction strategy}, which is the key technique of this paper and mitigates the local‐optima problem common in existing local reconstruction methods. 
Experimental results on both synthetic and real-world datasets demonstrate that LocalEscaper outperforms existing neural solvers, achieving remarkable results. 
\end{abstract}

\section{Introduction}

The Traveling Salesman Problem (TSP) is a classic combinatorial optimization (CO) problem with widespread applications in fields such as transportation~\cite{wang2021deep}, logistics~\cite{castaneda2022optimizing}, and circuit design~\cite{alkaya2013application}. Due to its NP-hard nature, finding the optimal solution for large-scale TSP instances remains a significant challenge. Over the past few decades, researchers have focused on mathematical programming and heuristic methods to find locally optimal solutions~\cite{applegate2009certification,lin1973effective,helsgaun2000effective}. However, these methods tend to be time-consuming, making them unsuitable for real-world, large-scale applications.

Recently, neural combinatorial optimization (NCO) methods have shown considerable promise in solving TSP~\cite{kwon2020pomo,jin2023pointerformer,drakulic2024bq}. 
Current NCO solvers typically rely on either supervised learning (SL)~\cite{joshi2019efficient,hottung2021learning,luo2023neural} or reinforcement learning (RL)~\cite{bello2016neural,kwon2020pomo,jin2023pointerformer}.  
SL-based methods require large datasets of labeled instances, where labels represent high-quality solutions typically obtained from exact solvers or heuristics. 
While SL-based methods are efficient in training, they depend on large amounts of labeled data, and generating high-quality labels for large-scale instances is computationally expensive.
On the other hand, RL-based methods do not rely on labeled data, learning to generate solutions through reward signals. However, 
RL methods face issues such as sparse rewards and a tendency to get stuck in local optima~\cite{bengio2021machine, min2024unsupervised}, which can reduce learning efficiency.

In addition to these challenges, scalability is a major concern for both SL- and RL-based NCO solvers, as most solvers plateau at problem sizes of around 500 to 1,000 nodes due to memory constraints. 
Moreover, most NCO solvers employ Transformer network architectures~\cite{vaswani2017attention}, which exhibit quadratic computational complexity as the problem scale increases. This makes it particularly difficult to train models capable of generating high-quality solutions for large-scale TSP.

To address the scalability issue, a number of NCO solvers have adopted divide-and-conquer strategies based on the Partial Optimization Metaheuristic Under Special Intensification Conditions (POPMUSIC) framework~\cite{taillard2019popmusic}, improving the solution quality by dividing the TSP tour into smaller, non-overlapping subsequences and reconstructing each subsequence independently~\cite{cheng2023select,luo2023neural,ye2024glop,zheng2024udc,luo2024boosting}. 
While these methods perform local search using subsequence reconstruction and reduce the complexity of solving large-scale problems, they are limited in their ability to escape local optima, as they only reconstruct the visiting order within subsequences while leaving the global relationships between subsequences unchanged.

Given these challenges, research on NCO methods for TSP faces three key obstacles: (1) Overcoming the limitations of SL and RL approaches, (2) escaping local optima induced by existing divide-and-conquer approaches, and (3) reducing computational costs. To address these challenges, we introduce LocalEscaper, a novel framework designed to enhance TSP solving. 

For the first challenge, SL often proves impractical for large‑scale problems, as obtaining ground‑truth (optimal‑solution) labels at scale is infeasible. Weakly supervised learning \cite{zhou2018brief} investigates how to learn effectively without access to high-quality labels. As a specific type of weak supervision, inaccurate supervision focuses on learning from noisy labeled data, where data-editing techniques \cite{brodley1999identifying,muhlenbach2004identifying} aim to identify and correct potential errors in the dataset to improve its quality. 
Inspired by this technique, we propose a weakly-supervised framework that combines SL and RL to leverage the strengths of both. 
Our method begins with SL using low-quality labels and progressively refines them through RL, enabling effective training without requiring high-quality labeled data. 
It is worth noting that our approach does not fall under the category of self-improvement learning techniques \cite{pirnay2024self,luo2024boosting}. Unlike such methods, which rely on a single model to refine labels and subsequently learn from them, our approach explicitly separates the processes of improvement and learning, employing different models for the two tasks. 

For the second challenge, we introduce a regional reconstruction strategy to overcome the limitations of existing divide-and-conquer approaches. This strategy is central to our framework and helps us better refine label quality. In contrast to traditional methods, which only adjust the visiting order within subsequences, our regional reconstruction method focuses on reconnecting edges within specific regions of the solution. This approach, coupled with an RL-based heuristic improver, enhances the solution quality by better escaping local optima during both training and inference stages.

For the third challenge, previous work \cite{fang2024invit} has improved neural network efficiency by pruning node information. Based on the fundamental observation that nodes closer to the current node are more likely to yield optimal selections \cite{fang2024invit}, selectively discarding distant nodes constitutes a reasonable strategy for reducing the model’s computational complexity. Motivated by this insight, we design a lightweight neural network architecture for our framework, which reduces both memory usage and computational overhead, thereby enabling scalability to large-scale TSP instances. 

Experimental results demonstrate that LocalEscaper achieves remarkable performance among existing NCO solvers on large-scale TSP instances ranging from 1,000 to 10,000 cities. In particular, LocalEscaper can handle large-scale instances with up to 100,000 cities quickly, while most existing NCO solvers either run out of memory or take too long.
The main contributions of our work are as follows: 
\begin{itemize}
\item We propose a novel weakly-supervised framework that enables training without high-quality labeled datasets, progressively refining label quality during the training process.
\item  We introduce a regional reconstruction approach, coupled with an RL-based improver, to enhance solution quality and effectively escape local optima during both training and inference stages. 
 \item  Experimental results demonstrate the superior performance of the proposed LocalEscaper over  existing NCO solvers on TSP instances ranging from 1,000 to 100,000 cities. 
\end{itemize}

\section{Related Work}

 In this section, we provide literature review from the following three perspectives on NCO Solvers. 
 
\subsection{SL-based NCO Solvers}
Supervised learning (SL)-based NCO solvers rely on large labeled datasets to train models.  
Early work in this domain was pioneered by \citet{vinyals2015pointer}, who proposed the Pointer Network.  Their approach used SL techniques to solve small-scale TSP problems, employing a recurrent neural network (RNN) with attention mechanisms to iteratively construct solutions. 
\citet{joshi2019efficient} advanced this by incorporating graph convolutional networks (GCNs) \cite{kipf2016semi} to predict the probability of each edge being part of the optimal solution. This helped the model better capture the underlying graph structure of the TSP. 
\citet{luo2023neural} introduced the Light
Encoder and Heavy Decoder (LEHD) model, which combines a lightweight encoder and a heavy decoder, making it particularly suitable for SL-based training. 
These SL-based methods have shown success for small-scale problems but still face challenges when scaling to larger TSP instances due to their reliance on large labeled datasets. 

\subsection{RL-based NCO Solvers}
Reinforcement learning (RL)-based NCO solvers, unlike SL-based methods, do not require labeled instances for training. 
Instead, they rely on reward signals to guide learning.  
\citet{kool2018attention} introduced a self-attention-based \cite{vaswani2017attention} NCO solver,  trained using RL, and demonstrated that it outperforms earlier SL-based methods in solving TSP instances. 
Building on this, \citet{kwon2020pomo} introduced the POMO solver, which generates multiple trajectories for a single TSP instance by starting from different nodes. 
During inference, POMO augments the input data with techniques like flipping and folding, enabling the model to produce diverse solutions and select the optimal one.  
\citet{jin2023pointerformer} improved POMO by incorporating a reversible residual network architecture to reduce memory consumption, enabling the model to scale to 500-node TSP instances with promising results. 

Despite these advancements, RL-based methods often struggle with issues such as getting stuck in local optima during training and the problem of sparse rewards \cite{bengio2021machine, min2024unsupervised}.
These challenges  hinder the ability of RL-based models to generalize to larger-scale problems. 

\subsection{Search-based NCO Solvers}
Search-based NCO solvers typically begin with an initial feasible solution and iteratively refine it to improve the outcome. Neural network models often play a direct or indirect role in guiding this refinement process. 
For instance, \citet{xin2021neurolkh} and \citet{zheng2023reinforced} integrated SL and RL into the classical heuristic Lin-Kernighan-Helsgaun (LKH) solver \cite{helsgaun2000effective,helsgaun2009general,helsgaun2017extension}, improving its efficiency by using learning-based methods to construct candidate node sets. 

Some search-based approaches leverage MCTS to enhance solution quality~\cite{fu2021generalize,qiu2022dimes,sun2023difusco,xia2024position}. These methods typically use GCNs to generate heatmaps that guide MCTS in finding better results. However, they are heavily reliant on MCTS, and constructing solutions greedily based solely on these heatmaps often leads to suboptimal performance. 
Additionally, MCTS is computationally expensive, making it difficult to scale for large TSP instances. 

Other search-based solvers use neural-based heuristics to directly refine solutions. 
For instance, \citet{cheng2023select} and \citet{ye2024glop} generated initial solutions using random insertion and applied subsequence reconstruction via RL-based models to enhance them. 
LEHD \cite{luo2023neural} also applied subsequence reconstruction on top of a constructive model to improve solutions. 
\citet{luo2024boosting} propose Self-Improved Training (SIT) solver, where the model is trained on a dataset with low-quality labels and then refines these labels using subsequence reconstruction. 

However, subsequence reconstruction methods have limited ability to modify the global relationships between nodes, often leading to the solver getting stuck in local optima. 
This limitation is a significant challenge for search-based NCO solvers. 

In contrast to self-improvement learning technique \cite{pirnay2024self,luo2024boosting}, we introduce the regional reconstruction algorithm during the process of improving solutions, which can address global relationships and effectively escape local optima. Additionally, our framework separates the solution improvement model from the solution construction model. Since each model is tailored to its specific task, they achieve better performance. Furthermore, the improvement model is trained via RL rather than SL, avoiding poor improvement performance caused by low-quality labels in the early training stage. In addition, the training of the constructive model and the refinement of labels can be carried out simultaneously.



\section{Preliminaries}
\subsection{The TSP Problem}
The focus of current NCO research is on the classic 2D Euclidean Traveling Salesman Problem (TSP). 
In this problem, we are given an undirected, fully connected graph $\mathcal{G}(\mathcal{V}, \mathcal{E})$ with $n$ nodes.  The node set $\mathcal{V}=\{v_i | 1\leq i \leq n\}$ represents the cities and the edge set $\mathcal{E}=\{e_{ij} | 1\leq i, j\leq n\}$ represents all possible connections between the cities. The cost, denoted as $\text{cost}(v_i, v_j)$, represents the Euclidean distance between nodes $v_i$ and $v_j$. 
The objective of the TSP is to find a Hamiltonian circuit (or tour) $\boldsymbol{\tau} = (\tau_1, \tau_2, \dots, \tau_n)$, which minimizes the total cost $L_{\text{total}}(\boldsymbol{\tau})$. The total cost is the sum of the Euclidean distances between consecutive nodes in the tour, as well as the distance from the last node back to the first, formulated as: 
\begin{equation}
{L_{\text{total}}(\boldsymbol{\tau})=\text{cost}(\tau_n, \tau_1)+\sum\limits_{i=1}^{n-1}{\text{cost}(\tau_i, \tau_{i+1})}},
\end{equation}
where $\tau_n$ is followed by $\tau_1$ to form the closed circuit.
  
\subsection{Subsequence Reconstruction Task}
In the subsequence reconstruction task, we are given a subsequence $\boldsymbol{\tau}^{\prime} = (\tau_1^{\prime}, \tau_2^{\prime}, \dots, \tau_{m}^{\prime})$ of length $m$, where $\boldsymbol{\tau}^{\prime} \subseteq \boldsymbol{\tau}$ and $m \leq n$. 
The goal is to reorder the intermediate nodes of the subsequence to form a new sequence $\boldsymbol{\tau}^{\prime\prime} = (\tau_1^{\prime\prime}, \tau_2^{\prime\prime}, \dots, \tau_{m}^{\prime\prime})$, while keeping the endpoints fixed: $\tau_1^{\prime\prime}=\tau_1^{\prime}$ and $\tau_m^{\prime\prime}=\tau_m^{\prime}$. 
The objective of subsequence reconstruction is to minimize the subsequence cost $L_{\text{sub}}(\boldsymbol{\tau}^{\prime\prime})$, which is given by:
\begin{equation}
{L_{\text{sub}}(\boldsymbol{\tau}^{\prime\prime})=\sum\limits_{i=1}^{m-1}{\text{cost}(\tau_i^{\prime\prime}, \tau_{i+1}^{\prime\prime})}},
\label{eq: SS obj.}
\end{equation}
where the goal is to reorder the subsequence to reduce the total cost, improving the overall tour quality.

\subsection{Regional Reconstruction Task}
The regional reconstruction task aims to improve the overall TSP solution by focusing on local regions within the tour. Given a tour $\boldsymbol{\tau}$, the corresponding set of edges $\mathcal{E}_{\boldsymbol{\tau}}=\{e_{\tau_{i},\tau_{i+1}} | 1\leq i \leq n-1\} \cup \{e_{\tau_{n},\tau_{1}}\}$ represents the set of connections in the tour. 

For a graph $\mathcal{G}$, we consider a 2D coordinate $c = (x, y)$ and define the set of $k$-nearest neighbors $\mathcal{V}_{c}$ in graph $\mathcal{G}$ as the nodes closest to $c$.
The edges in $\mathcal{E}_{\boldsymbol{\tau}}$ that have nodes from $\mathcal{V}_{c}$ as their predecessors are defined as $\mathcal{E}_{\boldsymbol{\tau}}^{\prime}=\{e_{\tau_{i},\tau_{i+1}} | \tau_{i} \in \mathcal{V}_{c}\}$. Specifically, define $\tau_{n+1}=\tau_{1}$ to denote the successor of $\tau_{n}$. 

The objective of the regional reconstruction task is to improve the current tour by modifying the edges of the regional set $\mathcal{E}_{\boldsymbol{\tau}}^{\prime}$.  
Specifically, we aim to find an edge set $\mathcal{E}^{\text{add}} \subset \mathcal{E}$ such that the updated edge set $\mathcal{E}_{\boldsymbol{\tau}}^{\text{new}} = \mathcal{E}^{\text{add}} \cup (\mathcal{E}_{\boldsymbol{\tau}} \setminus \mathcal{E}_{\boldsymbol{\tau}}^{\prime})$ forms a new Hamiltonian circuit $\boldsymbol{\tau}^\text{new}$. The goal is to minimize the total cost of the new tour $L_{\text{total}}(\boldsymbol{\tau}^\text{new})$, 
which directly corresponds to minimizing the cost of the newly added edges in $\mathcal{E}^{\text{add}}$, thus improving the overall solution quality.

\section{Methodology}
\begin{figure*}
\centering
\includegraphics[width=140mm]{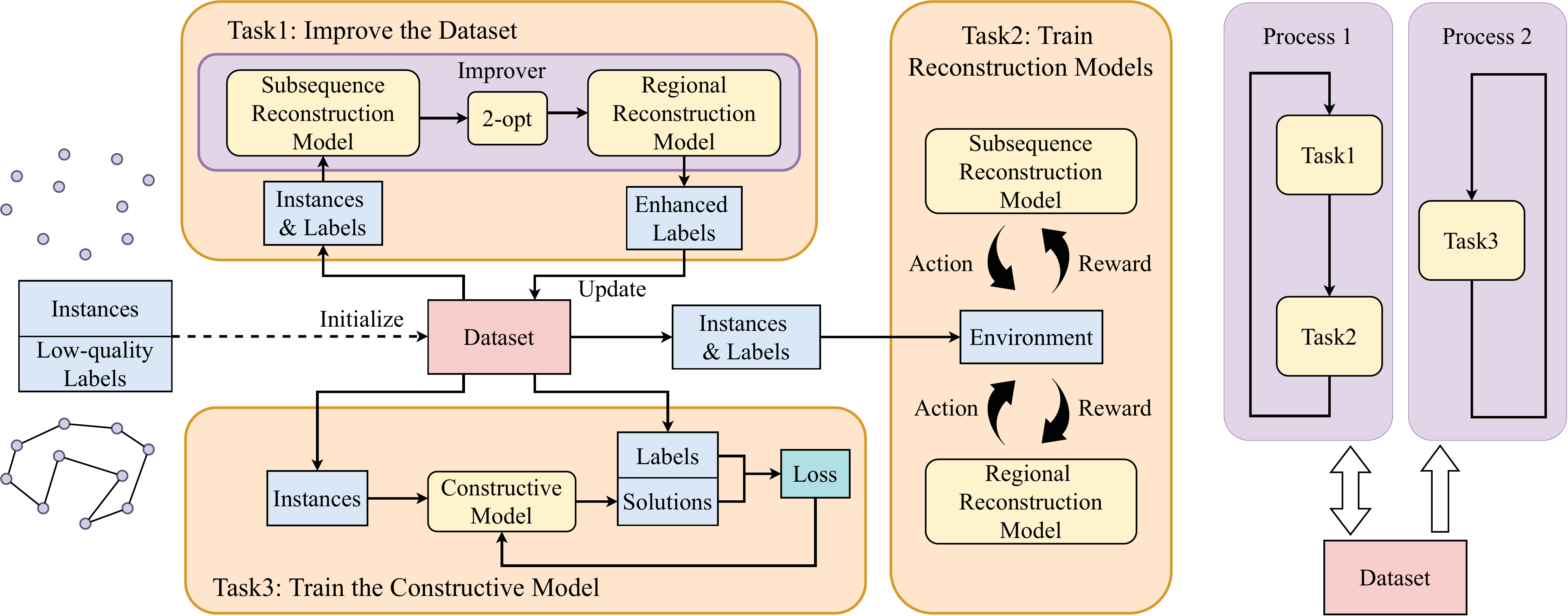}
\caption{Overview of the proposed framework, which includes a dataset and three tasks that can be executed in parallel. 
}
\label{fig:Framework}
\end{figure*}

In this section, we introduce a weakly-supervised learning framework to address the challenges of large-scale TSP discussed in the Introduction. An overview of the proposed framework is presented in Figure \ref{fig:Framework}. Our framework is designed to operate on a dataset consisting of three main tasks: 
Task 1 for dataset improvement, Task 2 for training reconstruction models, and Task 3 for training the constructive model. 
These tasks can be executed in parallel. For practical purposes, we adopt a sequential execution strategy where Task 1 and Task 2 alternate, while Task 3 runs continuously. 

\subsection{Weakly-supervised Learning Framework}
We define the dataset as $\mathcal{D}^t(\boldsymbol{\mathcal{G}}, \boldsymbol{\mathcal{T}}^t)$, where 
$\boldsymbol{\mathcal{G}}$ 
represents the set of TSP graphs, 
$\boldsymbol{\mathcal{T}}^t$ 
corresponds to the set of tour labels, with  $t \in \{0,1,\dots,T\}$ indicating the iteration counter. 
The initial label set, $\boldsymbol{\mathcal{T}}^0$, is generated using a random insertion algorithm based on the graph set $\boldsymbol{\mathcal{G}}$.

We employ a weakly-supervised learning approach, where the dataset $\mathcal{D}$ is used to train a constructive model parameterized by $\boldsymbol{\theta}$. This model generates TSP solutions by sequentially selecting the next node to visit. 

\textbf{Task1: Improve the Dataset}.
To improve the label quality of the dataset, we introduce an improver $\mathcal{I}$, which is independent of the model $\boldsymbol{\theta}$. 
The label improvement process is defined as $\mathcal{D}^{t+1} = \mathcal{I}(\mathcal{D}^t)$, and we impose the constraint that the total loss for the updated labels satisfies: 
$L_{\text{total}}(\boldsymbol{\tau}^{t+1}) \leq L_{\text{total}}(\boldsymbol{\tau}^{t})$, $\forall \boldsymbol{\tau}^{t} \in \boldsymbol{\mathcal{T}}^{t}$. 

\textbf{Task2: Train Reconstruction Models}.
To enhance the performance of the improver $\mathcal{I}$, we establish subsequence and regional reconstruction tasks on the updated dataset $\mathcal{D}^{t+1}$ as environments to train the corresponding reconstruction models using reinforcement learning.

\textbf{Task3: Train  the Constructive Model}.
The constructive model $\boldsymbol{\theta}$ follows the "learning to construct partial solutions" paradigm proposed in \citet{luo2023neural}.
At each training step, we sample a subsequence $\boldsymbol{\tau}^{\prime}$ from a given label $\boldsymbol{\tau}$, where $\boldsymbol{\tau}^\prime$ provides a binary indication $b_i$ (1 for selection, 0 for non-selection) for each node $v_i$ in $\boldsymbol{\tau}^\prime$. 
The model then predicts the selection probability $p_i$ for each unvisited node $v_i$. 
The objective function is the cross-entropy loss:
${\mathcal{L}(\boldsymbol{\theta}) = -\sum_{i=1}^{u-1}{b_i\text{log}(p_i)}}$,
where $u$ is the number of unvisited nodes in the subsequence.


Appendix \ref{Appendix-Pseudocode} provides the pseudocode of the framework. Next, we provide a detailed description of the improver.

\subsection{Improver for Escaping Local Optima}
A crucial component of our framework is the improver $\mathcal{I}$, designed to help escape local optima by refining solutions through regional reconstruction. 
The improver operates both during training (to improve the label quality of the dataset) and during inference (to enhance the quality of the generated solutions). 
The improver follows a three-step process: Subsequence Reconstruction, 2-opt, and Regional Reconstruction.

\textbf{Subsequence Reconstruction}: Building upon the approach in \citet{ye2024glop}, we randomly decompose a TSP solution of length $n$ into $\lfloor \frac{n}{m} \rfloor$ subsequences, each of length $m$. 
Each subsequence is then reconstructed using a neural model parameterized by $\boldsymbol{\psi}$. 
The model replaces the original subsequence $\boldsymbol{\tau}^{\prime}$ with a newly reconstructed subsequence \textbf{$\boldsymbol{\tau}^{\prime\prime}$},   if $L_{\text{sub}}(\boldsymbol{\tau}^{\prime\prime}) \leq L_{\text{sub}}(\boldsymbol{\tau}^{\prime})$.

We employ a non-autoregressive neural model for subsequence reconstruction task. Specifically, the encoder generates a heatmap 
of edge selection probabilities in a single pass, and the decoder constructs new subsequences iteratively based on 
the heatmap. 
The training and inference of the model follow the multiple trajectories approach~\cite{kwon2020pomo}. 
Non-autoregressive models offer fast construction speeds. When the subsequence length $m$ is small (e.g., less than 100), the drawback of such coarse construction~\cite{ye2024glop} is minimal. 
The Appendix \ref{Appendix-Subsequence Reconstruction} provides specific details of the subsequence reconstruction approach.

\textbf{2-opt}: We apply the classical 2-opt algorithm~\cite{lin1973effective}, which iteratively checks all possible  edge swaps to improve the solution. 
If a swap leads to a shorter solution, it is executed, and the process continues  until no further improvements are possible. This helps fine-tune the solution by removing suboptimal edges, especially in the early stages of improvement.

\textbf{Regional Reconstruction}: 
According to the definition of the regional reconstruction task in the preliminaries section, we remove a set of edges $\mathcal{E}_{\boldsymbol{\tau}}^{\prime}$ from the original TSP solution $\mathcal{E}_{\boldsymbol{\tau}}$.  Let $|\mathcal{E}_{\boldsymbol{\tau}}^{\prime}|=k$, the remaining edges $\mathcal{E}_{\boldsymbol{\tau}} \setminus \mathcal{E}_{\boldsymbol{\tau}}^{\prime}$ can compose $k$ subsequences 
$\mathcal{S}=\{ \boldsymbol{\tau}_{1}^{\prime}, \boldsymbol{\tau}_{2}^{\prime}, \dots, \boldsymbol{\tau}_{k}^{\prime} \}$
, where each subsequence is independent of the direction of traversal.

To ensure that the reconstructed solution is a valid Hamiltonian circuit, we transform the reconstruction process into a permutation problem. We permute the subsequences in $\mathcal{S}$ to form a subsequence list $\mathcal{X}=(\boldsymbol{\eta}_{1}, \boldsymbol{\eta}_{2}, \dots, \boldsymbol{\eta}_{k})$. 
During the permutation, we have the option to reverse the direction of any subsequence.  
For instance, for a subsequence $\boldsymbol{\tau}^{\prime} = (\tau_1^{\prime}, \tau_2^{\prime}, \dots, \tau_{m}^{\prime})$, its reverse is denoted as $\neg \boldsymbol{\tau}^{\prime} = (\tau_{m}^{\prime}, \tau_{m-1}^{\prime}, \dots, \tau_{1}^{\prime})$. 

Once the subsequences are permuted and reversed as necessary, they are concatenated to form a new Hamiltonian circuit $\boldsymbol{\tau}^{\text{new}}$. This reconstructed solution provides a refined path that potentially escapes local optima by incorporating a broader set of possible configurations.

We model the permutation task as a Markov Decision Process (MDP) to optimally select subsequences and reconstruct the final solution. 
In this framework, 
the subsequence list $\mathcal{X}$ is gradually built by appending subsequences from a set of available subsequences $\mathcal{C}$.

The MDP  formulation consists of the following elements: states, actions, transition, and rewards. 

\textit{States}: The state is characterized by the current subsequence list $\mathcal{X}$ and the set of available subsequences $\mathcal{C}$.
Initially, $\mathcal{X}$ is an empty list, and $\mathcal{C}$ contains the subsequences as well as their reverses:
$\mathcal{C}=\{\boldsymbol{\tau}_{1}^{\prime}, \neg\boldsymbol{\tau}_{1}^{\prime}, \boldsymbol{\tau}_{2}^{\prime}, \neg\boldsymbol{\tau}_{2}^{\prime},\dots,\boldsymbol{\tau}_{k}^{\prime}, \neg\boldsymbol{\tau}_{k}^{\prime}\}$. 

\textit{Actions}: An action is the selection of a subsequence $\boldsymbol{\tau}_{\text{act}}$ from the available set $\mathcal{C}$ to be appended to the subsequence list $\mathcal{X}$.

\textit{Transition}: After performing an action, the subsequence $\boldsymbol{\tau}_{\text{act}}$ is appended to $\mathcal{X}$, and both $\boldsymbol{\tau}_{\text{act}}$ and its reverse $\neg\boldsymbol{\tau}_{\text{act}}^{\prime}$ are removed from $\mathcal{C}$.
The process continues until all subsequences are appended, and  $\mathcal{C}$ becomes empty, signaling the completion of the reconstruction. 

\textit{Rewards}: The reward function is designed to evaluate the quality of the reconstructed solution. 
Once the reconstruction is complete, the total reward is defined as  the negative of the solution's cost, which is the sum of the edge costs between adjacent subsequences: 
\begin{equation}
{R(\mathcal{X})=-\text{cost}(\eta_{1}^{\text{first}}, \eta_{k}^{\text{last}})-\sum\limits_{i=1}^{k-1}\text{cost}(\eta_{i}^{\text{last}}, \eta_{i+1}^{\text{first}})},
\label{eq:Reward}
\end{equation}
where $\eta_{i}^{\text{first}}$ and $\eta_{i}^{\text{last}}$ represent the first and last nodes of subsequence $\boldsymbol{\eta}_i$, respectively. 
The goal is to minimize this total cost by carefully selecting subsequences during the reconstruction process. 

We employ the REINFORCE algorithm~\cite{williams1992simple} to train a non-autoregressive policy model $\boldsymbol{\phi}$ to select subsequences during regional reconstruction.  
The policy model is trained by sampling multiple trajectory lists $\boldsymbol{\mathcal{X}}=\{\mathcal{X}^{1}, \mathcal{X}^{2}, \dots, \mathcal{X}^{N} \}$, where each $\mathcal{X}^i$ represents a possible subsequence list generated by the policy. 

The gradient of the policy network is approximated as: 
\begin{equation}
\begin{aligned}
\triangledown_{\boldsymbol{\phi}}J(\boldsymbol{\phi}) &\approx \frac{1}{N}\sum\limits_{i=1}^{N} \frac{R(\mathcal{X}^{i}) - \mu(\boldsymbol{\mathcal{X}})}{\delta(\boldsymbol{\mathcal{X}})} \triangledown_{\boldsymbol{\phi}} \text{log} ~ p_{\boldsymbol{\phi}}(\mathcal{X}^{i}| \mathcal{S}),
\label{eq:Regional Reconstruction Gradient}
\end{aligned}
\end{equation}
Here, $p_{\boldsymbol{\phi}}(\mathcal{X}^{i}| \mathcal{S})$ is the probability of the policy model $\boldsymbol{\phi}$ constructing the solution $\mathcal{X}^{i}$ given the regional reconstruction problem $\mathcal{S}$. 
Symbols $\mu(\boldsymbol{\mathcal{X}})$ and $\delta(\boldsymbol{\mathcal{X}})$ denote the mean and standard deviation of rewards $R(\boldsymbol{\mathcal{X}})$ for trajectories $\boldsymbol{\mathcal{X}}$ respectively.

The training process aims to optimize the policy so that subsequences are selected in a way that minimizes the total cost of the final reconstructed solution. During this process, the same coordinate transformation used in subsequence reconstruction is applied, and the new solution replaces the original one if its cost is lower. More details can be found in Appendix \ref{Appendix-Regional Reconstruction}. 

\subsection{Lightweight Constructive Model for TSP}
The attention mechanism~\cite{vaswani2017attention} is computationally expensive when applied to large-scale problems such as the TSP
~\cite{kwon2020pomo,luo2023neural}. 
Previous work \cite{fang2024invit} selects key nodes to reduce the number of inputs to the model, thereby enhancing computational efficiency. 
Inspired by this insight, our constructive model adopts a more lightweight architecture.

In the encoder, a 2-layer perceptron maps the coordinates of nodes into encoding embeddings. During the decoding phase, we prune a part of nodes to reduce the decoder’s computational overhead. For the last visited node $\tau_\text{last}$, we select the $k_d$ nearest unvisited nodes of $\tau_\text{last}$ as the candidate set $\mathbb{A}^\text{p}$. 
If fewer than $k_d$ nodes remain unvisited, we include all remaining nodes in the candidate set. 
Then we feed candidates $\mathbb{A}^\text{p}$, the last visited node $\tau_\text{last}$ and the first visited node into the decoder. 
Finally, the model calculates the selection probabilities of each node in $\mathbb{A}^\text{p}$ through multiple attention layers and a softmax function. 
Since each attention layer processes at most $k_d+2$ nodes, its computational complexity does not increase as the overall problem scale grows. 
Detailed information is provided in Appendix \ref{Appendix-Lightweight Model} due to the space limitation.

\section{Experiments}
\begin{table*}[htb]
\caption{Comparison of results on large-scale TSP instances. RL, SL, G, S, NH and MCTS denote Reinforcement Learning, Supervised Learning, Greedy Decoding, Sampling Decoding, Neural-based Heuristics and Monte Carlo Tree Search, respectively. * indicates the baseline used for calculating the performance gap.}
\setlength{\tabcolsep}{0.5mm}
\centering
\small
\begin{tabular}{l c
| cc cc cc
| cc cc cc }
\toprule
\multicolumn{2}{c|}{\textbf{Distribution}}
& \multicolumn{6}{c|}{\textbf{Uniform}} & \multicolumn{6}{c}{\textbf{Clusted}} \\
\midrule
\multirow{2}{*}{Method} & \multirow{2}{*}{Type} & \multicolumn{2}{c}{TSP1K} & \multicolumn{2}{c}{TSP5K} & \multicolumn{2}{c|}{TSP10K}
& \multicolumn{2}{c}{TSP1K} & \multicolumn{2}{c}{TSP5K} & \multicolumn{2}{c}{TSP10K} \\
& & Gap($\%$) & Time & Gap($\%$) & Time & Gap($\%$) & Time
& Gap($\%$) & Time & Gap($\%$) & Time & Gap($\%$) & Time \\
\midrule

Concorde       & Exact     & 0.00* & 2.21h &   \multicolumn{2}{c}{N/A}    &  \multicolumn{2}{c|}{N/A}     
                           & 0.00* & 11.30h &  \multicolumn{2}{c}{N/A}      &  \multicolumn{2}{c}{N/A} \\
LKH-3          & Heuristic& 0.00 & 4.41h & 0.00*& 2.62h  & 0.00* & 6h    
                           & 0.00 & 1.93h  & 0.00* & 0.56h  & 0.00* & 1.76h \\
                           \midrule
POMO        & RL+S     & 40.61 & 8.60m &  \multicolumn{2}{c}{OOM}      &  \multicolumn{2}{c|}{OOM}     
                           & 48.68 & 13.67m & \multicolumn{2}{c}{OOM}      &  \multicolumn{2}{c}{OOM} \\
Pointerformer  & RL+S   & 7.27  & 6.33m &  \multicolumn{2}{c}{OOM}      &  \multicolumn{2}{c|}{OOM}     
                           & 32.42 & 12.43m &  \multicolumn{2}{c}{OOM}      &  \multicolumn{2}{c}{OOM} \\
INViT-3V       & RL+G     & 5.84  & 0.62h & 6.74 & 29.93m & 7.10 & 0.93h  
                           & 7.48  & 0.82h  & \textbf{9.13} & 31.11m & \textbf{9.45} & 1.46h \\
BQ-NCO        &  SL+G        & 4.20  & 10.05m&12.86 & 2.67h  &21.17 & 22.12h 
                           & 9.61  & 15.85m &26.92 & 3.32h  &43.91 & 27.52h \\
H-TSP          & RL+S     & 6.66  & 55.63s& 9.47 & 29.24s & 9.43 & 58.24s 
                           & 8.33  & 80.80s & 9.69 & 41.66s & 9.78 & 1.28m \\
\textbf{Ours} greedy    & SL+G     & \textbf{2.67}  & \textbf{17s}   & \textbf{3.28} & \textbf{21s}    & \textbf{4.66} & \textbf{43s}   
                           & \textbf{4.28}  & \textbf{26s}    &10.32 & \textbf{23s}    &18.86 & \textbf{47s} \\
                           \midrule
GLOP          & RL+NH    & 3.11  & 3.30m & 4.69 & \textbf{47s}    & 4.89 & \textbf{1.33m} 
                           & 3.65  & 5.49m  & 5.32 & \textbf{58.40s} & 5.26 & \textbf{1.90m} \\
Attn-GCN     & SL+MCTS  & 2.47  & 4.3m  & 3.32 & 15.02m & 3.56 & 42.68m
                           & 2.30  & 6.29m  & 3.24 & 18.42m & 3.39 & 0.96h \\
DIFUSCO  & SL+MCTS  & 1.12  & 13.73m& 2.53 & 22.95m & 2.56 & 1.07h 
                           & 1.51  & 29.39m & 2.51 & 22.50m & 3.24 & 0.91h \\
LEHD RRC100   & SL+NH    & 1.73  & 42.73m&10.70 & 13.60h & \multicolumn{2}{c|}{N/A}    
                           & 4.64  & 1.30h  &16.90 & 18.57h & \multicolumn{2}{c}{N/A} \\
SIT PRC 100   & SL+NH    & \textbf{0.48}  & 20.52m& 1.69 & 15.30m & 2.11 & 31.49m
                           & \textbf{0.80}  & 32.62m & 2.72 & 18.31m & 4.05 & 38.57m \\
\textbf{Ours} Rec100   & SL+RL+NH & 0.94  & \textbf{1.83m} & 1.41 & 1.08m  & 1.83 & 3.07m 
                           & 1.20  & \textbf{2.85m}  & 2.24 & 1.60m  & 2.84 & 7.48m \\
\textbf{Ours} Rec500   & SL+RL+NH & 0.75  & 7.95m & \textbf{1.22} & 3.77m  & \textbf{1.62} & 8.45m
                           & 0.94  & 12.43m & \textbf{2.00} & 4.82m  & \textbf{2.60} & 14.28m \\
\midrule

\multicolumn{2}{c|}{\textbf{Distribution}}
& \multicolumn{6}{c|}{\textbf{Explosion}} & \multicolumn{6}{c}{\textbf{Implosion}} \\
\midrule
\multirow{2}{*}{Method} & \multirow{2}{*}{Type} & \multicolumn{2}{c}{TSP1K} & \multicolumn{2}{c}{TSP5K} & \multicolumn{2}{c|}{TSP10K}
& \multicolumn{2}{c}{TSP1K} & \multicolumn{2}{c}{TSP5K} & \multicolumn{2}{c}{TSP10K} \\
& & Gap($\%$) & Time & Gap($\%$) & Time & Gap($\%$) & Time
& Gap($\%$) & Time & Gap($\%$) & Time & Gap($\%$) & Time \\
\midrule
Concorde       & Exact    & 0.00* & 11.80h &  \multicolumn{2}{c}{N/A}    & \multicolumn{2}{c|}{N/A}    
                           & 0.00* & 22.02h & \multicolumn{2}{c}{N/A}    & \multicolumn{2}{c}{N/A} \\
LKH-3          & Heuristic& 0.00 & 0.70h  & 0.00* & 27.33m & 0.00* & 1.37h  
                           & 0.00 & 6.03h  & 0.00* & 2.18h & 0.00* & 8.41h \\
                           \midrule
POMO       & RL+S     & 47.83 & 13.66m &  \multicolumn{2}{c}{OOM}   &  \multicolumn{2}{c|}{OOM}     
                           & 40.36 & 13.72m &  \multicolumn{2}{c}{OOM}    &  \multicolumn{2}{c}{OOM} \\
Pointerformer  & RL+S   & 24.18 & 12.40m &  \multicolumn{2}{c}{OOM}    &  \multicolumn{2}{c|}{OOM}     
                           & 9.40  & 12.45m &  \multicolumn{2}{c}{OOM}    &  \multicolumn{2}{c}{OOM} \\
INViT-3V       & RL+G     & 9.20  & 48.31m & 11.50 & 30.81m & 10.49 & 1.54h  
                           & 6.97  & 53.63m & 7.96 & 33.44m & \textbf{7.83} & 1.53h \\
BQ-NCO        &  SL+G        & 7.45  & 15.69m & 27.76 & 3.35h  & 53.02 & 27.35h 
                           & 5.28  & 15.61m & 15.82 & 3.33h & 26.98 & 27.44h \\
H-TSP          & RL+S     & 8.26  & 1.49m  & 10.03 & 41.05s & \textbf{10.37} & 1.39m 
                           & 7.37  & 1.34m  & 9.65  & 40.43s & 10.09 & 1.35m \\
\textbf{Ours} greedy    & SL+G     & \textbf{5.03}  & \textbf{26s}    & \textbf{8.7}   & \textbf{23s}   & 21.51 & \textbf{46s}   
                           & \textbf{4.98}  & \textbf{26s}    & \textbf{5.87}  & \textbf{23s}    & 8.4   & \textbf{46s} \\
                           \midrule
GLOP          & RL+NH    & 2.96  & 5.37m  & 5.06  & \textbf{57.27s} & 5.32  & \textbf{1.93m}  
                           & 2.85  & 5.39m  & 4.81  & \textbf{57.23s} & 4.94  & \textbf{1.91m} \\
Attn-GCN     & SL+MCTS  & 2.15  & 6.05m  & 3.56  & 18.59m & 4.56  & 58.03m 
                           & 2.31  & 6.14m  & 3.92  & 18.56m & 4.34  & 56.47m \\
DIFUSCO  & SL+MCTS  & 1.34  & 29.39m & 3.22  & 22.53m & 4.43  & 53.44m 
                           & 1.17  & 29.54m & 2.99  & 22.60m & 3.49  & 53.57m \\
LEHD RRC100   & SL+NH    & 2.84  & 1.34h  & 15.44 & 17.14h & \multicolumn{2}{c|}{N/A}     
                           & 1.71  & 1.34h  & 11.47 & 17.11h & \multicolumn{2}{c}{N/A} \\
SIT PRC 100   & SL+NH    & \textbf{0.71}  & 33.19m & 2.28  & 18.42m & 3.47  & 38.78m 
                           & \textbf{0.67}  & 32.96m & 1.92  & 18.37m & 2.36  & 38.70m \\
\textbf{Ours} Rec100   & SL+RL+NH & 1.15  & \textbf{2.95m}  & 2.12  & 1.53m  & 2.86  & 6.85m  
                           & 1.07  & \textbf{2.83m}  & 1.55  & 1.48m  & 1.99  & 5.12m \\
\textbf{Ours} Rec500   & SL+RL+NH & 0.90  & 12.82m & \textbf{1.83}  & 4.90m  & \textbf{2.58}  & 13.43m
                           & 0.83  & 12.13m & \textbf{1.33}  & 4.83m  & \textbf{1.75}  & 11.67m \\
\bottomrule
\end{tabular}
\label{Table: TSP random}
\end{table*}

\begin{figure*}[t]
    \centering  
	\subfigbottomskip=2pt 
	\subfigcapskip=-5pt 
	\subfigure[Gap vs. Iterations]{
		\includegraphics[width=0.60\columnwidth]{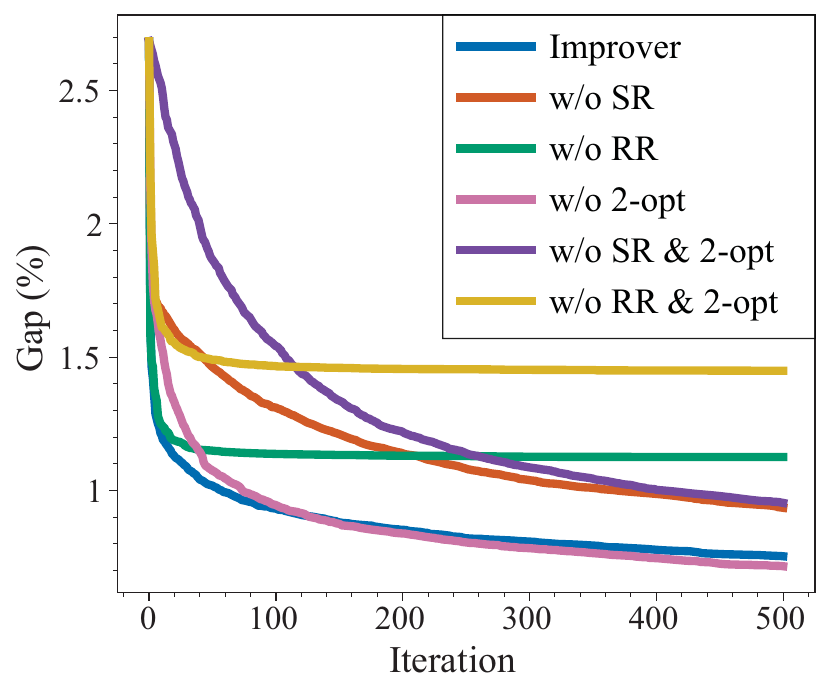}}
    \subfigure[Time vs. Iterations]{
		\includegraphics[width=0.60\columnwidth]{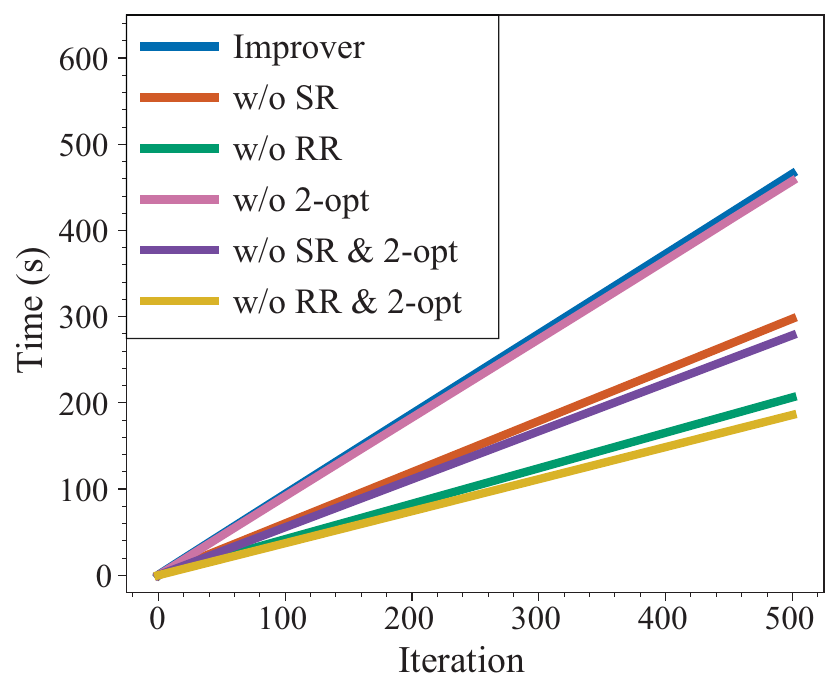}}
    \subfigure[Gap vs. Time]{
		\includegraphics[width=0.60\columnwidth]{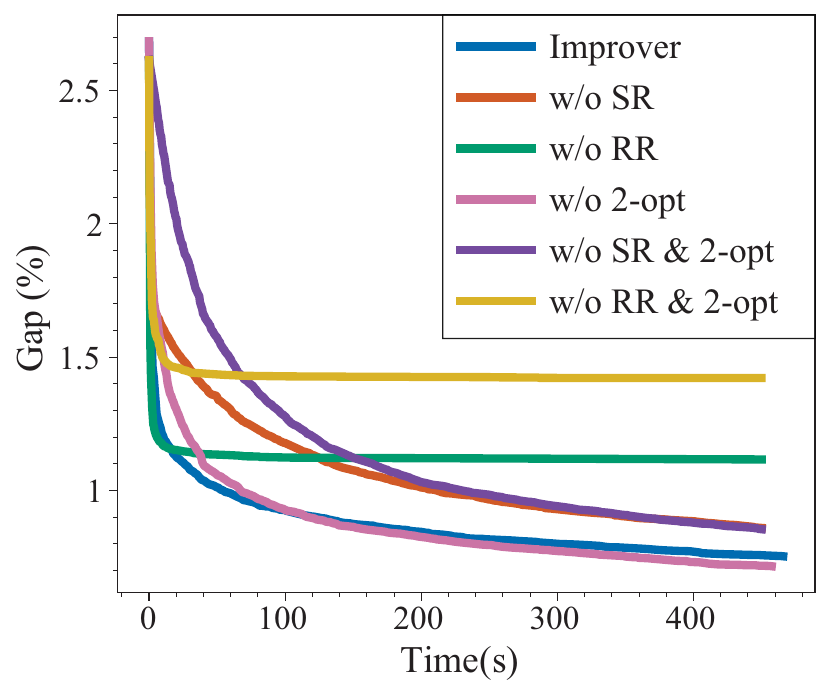}}
    \caption{Ablation study on the three components of the improver in uniform TSP1K. SR and RR denote Subsequence Reconstruction and Regional Reconstruction  respectively. 
    }
    \label{fig:Decomposition}
\vspace{-1em}
\end{figure*}

\begin{table*}[ht]
\centering
\small
\caption{Comparison of results on TSPLIB benchmark instances. TSPLIB51$\sim$1000 contains 48 instances, TSPLIB1001$\sim$5000 contains 22 instances, and TSPLIB5001+ contains 7 instances.}
\setlength{\tabcolsep}{1mm}
\begin{tabular}{cc|ccr|ccr|ccr}
\toprule
\multirow{2}{*}{Method} & \multirow{2}{*}{Type}
& \multicolumn{3}{c|}{TSPLIB51 $\sim$ 1000} & \multicolumn{3}{c|}{TSPLIB1001 $\sim$ 5000} & \multicolumn{3}{c}{TSPLIB5001+} \\
&&Obj.&Gap&Time&Obj.&Gap&Time&Obj.&Gap&Time \\
\midrule

OPT & N/A & 34108.69 & 0.00\% & N/A & 158350.18 & 0.00\% & N/A & 3530775.57 & 0.00\% & N/A \\


GLOP & RL+NH & 34542.40 & 1.23\% & 2.46m & 167577.95 & 5.96\% & 1.46m & 3756502.71 & 7.87\% & 1.75m \\

Att-GCN & SL+MCTS & 34370.29 & 0.92\% & 19.83m & 163793.09 & 7.02\% & 1.18h &\multicolumn{3}{c}{OOM} \\

DIFUSCO & SL+MCTS & 34242.58 & 0.43\% & 21.78m & 162367.82 & 3.15\% & 1.24h & 3864066.71 & 5.21\% & 2.64h \\


LEHD RRC100 & SL+NH & 34629.54 & 1.41\% & 1.07h  &173601.50 &11.10\% &5.24h &\multicolumn{3}{c}{OOM} \\

SIT PRC100 & SL+NH &34292.50 &0.37\% &1.10h  &160159.50 &1.50\% &2.25h  &3794651.00 &4.63\% &46.43m \\

\textbf{LocalEscaper} Rec500 & SL+RL+NH &34192.60 &\textbf{0.33\%} &28.92m  &160394.59 &\textbf{1.44\%} &14.27m  &3645591.00 &\textbf{2.74\%} &10.66m \\

\bottomrule
\end{tabular}
\label{Table: TSPLIB}
\end{table*}

\subsection{Experimental Setup}
\textbf{Problem Settings}. We adopt the standard data generation process from the previous work~\cite{kool2018attention}. 
The training set consists of 8192 instances, each with 1000 nodes randomly sampled from a uniform distribution. 
The test set includes both synthetic data and real-world data from TSPLIB~\cite{reinelt1991tsplib}. 
The synthetic data includes instances with three problem sizes: 1000, 5000, and 10000 nodes, referred to as TSP1K, TSP5K, and TSP10K, respectively. 
For each problem size, we prepare instances from four distributions: uniform, clustered, explosion, and implosion. 
The uniform TSP1K and TSP10K test sets use the same instances as in \citet{fu2021generalize}, with 128 and 16 instances, respectively. The TSP5K test set is generated similarly, with 16 instances. The other three distributions test sets use the same instances are the same with \citet{fang2024invit}, with 200 instances for TSP1K and 20 instances for the other size. 
For subsequence reconstruction problems, each subsequence length is set to $m=100$. For regional reconstruction problems, the center point $c$ is sampled from the $[0,1]^2$ space, and the number of removed edges is set to $k=60$. 

\textbf{Model Settings}. The constructive model $\boldsymbol{\theta}$ consists of  $L=6$ decoding layers. Both the subsequence reconstruction model $\boldsymbol{\psi}$ and the regional reconstruction model $\boldsymbol{\phi}$ have 6 encoding layers. The maximum size of candidate set is $k_d=200$. 
We evaluate LocalEscaper in three modes: 
(1) \textbf{LocalEscaper greedy}: A greedy construction using the constructive model; 
(2) \textbf{LocalEscaper Rec100}: After greedy construction, the solution is refined with 100 improvement iterations by our improver; 
(3) \textbf{LocalEscaper Rec500}: Similar to Rec100, but with up to 500 improvement iterations. 

\textbf{Training}. 
We train the constructive model starting with shorter subsequences and progressively increasing their length as training progresses. 
Specifically, the subsequence length $|\boldsymbol{\tau}^{\prime}|$ is sampled from the range $[50,100]$ for each batch. 
After each epoch, the upper bounds of the sampling range are increased by $5$, continuing until it reaches $1000$ nodes. 
The optimizer is Adam~\cite{kingma2014adam}, with an initial learning rate of 1e-4 and a decay rate of $0.98$ for the first 200 epochs. Both the reconstruction models $\boldsymbol{\psi}$ and $\boldsymbol{\phi}$ are trained with the same optimizer and learning rate. We summarize the experimental settings in Appendix \ref{Appendix-Training setting}. 

\textbf{Baselines}. We compare our method with 
(1) \textbf{Classical solvers}: 
Concorde~\cite{applegate2009certification} and 
LKH-3~\cite{helsgaun2017extension}; 
(2) \textbf{Constructive solver}s: POMO~\cite{kwon2020pomo}, Pointerformer~\cite{jin2023pointerformer}, BQ-NCO\cite{drakulic2024bq}, H-TSP~\cite{pan2023h}, INViT-3V\cite{fang2024invit}; (3) \textbf{Search-based solvers}: Att-GCN~\cite{fu2021generalize}, DIFUSCO~\cite{sun2023difusco}, GLOP~\cite{ye2024glop}, LEHD RRC100~\cite{luo2023neural} and SIT~\cite{luo2024boosting}. 

\textbf{Metrics and Inference}.
We evaluate the methods based on two metrics: average optimality gap (Gap) and total run-time (Time). 
Since classical solvers are run on a single CPU, their run-time should not be directly compared with methods running on a GPU. 
All neural solvers are run on an RTX 3090 GPU for inference. We set a maximum inference time of 30 hours per problem size, and any method exceeding this time limit is marked as "N/A". 

\subsection{Results and Analysis}
The main experimental results on the instances from four distribution types are presented in Table \ref{Table: TSP random}. 
In comparison with constructive solvers, our LocalEscaper approach demonstrates fast inference speed under greedy decoding and achieves the lowest gap across all three instance sizes in uniform instances. Under the other distributions, our method also demonstrates advanced performance and efficiency. 

Compared to search-based solvers, LocalEscaper achieves the lowest gap in TSP5K and TSP10K across all distributions. 
This demonstrates LocalEscaper’s superior performance on large-scale problems. Moreover, the results collectively indicate that our method exhibits outstanding generalization and efficiency. 

Experimental results on TSPLIB are shown in Table \ref{Table: TSPLIB}. We categorize instances into three groups based on their size: $51\sim1000$ nodes, $1001\sim5000$ nodes, and over 5000 nodes. For each group, we solve all instances sequentially. 
Emphasizing the optimality gap, we compare LocalEscaper with other search-based methods. LocalEscaper achieves the lowest gap across all instance categories. More details can be found
in Appendix \ref{Appendix-TSPLIB}. 


\subsection{Ablation Study} 

\textbf{Effect of the Improver's Components}. 
We conducted an ablation study on uniform TSP1K to investigate the impact of each component of the improver.  
Figure \ref{fig:Decomposition} shows the solution improvement process over 500 iterations, with one or two components of the improver removed. Removing 2-opt exerts negligible influence on the results after convergence. Removing either regional reconstruction or both regional reconstruction and 2-opt significantly reduces performance, indicating that relying solely on subsequence reconstruction leads to a fast convergence to low-quality local optima. 



\begin{table}[ht]
\caption{Effect of regional reconstruction scale on performance over 100 improvement iterations.}
\centering
\small
\setlength{\tabcolsep}{0.8mm}
\begin{tabular}{c|cc|cc|cc}
\toprule
\multirow{2}{*}{Scale} 
& \multicolumn{2}{c|}{TSP1K} & \multicolumn{2}{c|}{TSP5K} & \multicolumn{2}{c}{TSP10K}\\
&Gap&Time &Gap&Time &Gap&Time\\
\midrule

$k=30$ &1.00\% &1.12m  &1.50\% &0.72m  &1.89\% &2.26m \\

$k=60$ &0.94\% &1.54m  &1.41\% &0.73m  &1.83\% &2.35m \\

$k=90$ &0.87\% &2.07m  &1.40\% &0.83m  &1.81\% &2.5m \\
\bottomrule
\end{tabular}
\label{Table: RR100}
\end{table}

\begin{table}[ht]
\caption{Effect of regional reconstruction scale on performance over 500 improvement iterations.}
\small
\centering
\setlength{\tabcolsep}{0.8mm}
\begin{tabular}{c|cc|cc|cc}
\toprule
\multirow{2}{*}{Scale} 
& \multicolumn{2}{c|}{TSP1K} & \multicolumn{2}{c|}{TSP5K} & \multicolumn{2}{c}{TSP10K}\\
&Gap&Time &Gap&Time &Gap&Time\\
\midrule

$k=30$ &0.86\% &5.56m  &1.34\% &3.10m  &1.70\% &7.25m \\

$k=60$ &0.75\% &7.66m  &1.22\% &3.42m  &1.62\% &7.73m \\

$k=90$ &0.65\% &10.28m  &1.19\% &3.8m  &1.54\% &8.32m \\
\bottomrule
\end{tabular}
\label{Table: RR500}
\end{table}
\textbf{Effect of Regional Reconstruction Scale}
We investigated the impact of the number of removed edges $k$ in the regional reconstruction process on the uniform test sets. We evaluated three settings: $k=30$, 60 and 90. For each configuration, we retrained the regional reconstruction model $\boldsymbol{\phi}$ on the training set $\mathcal{D}^{1200}$, which was obtained after 1200 improvement iterations, using a learning rate of 1e-4 and training for 1000 epochs. 
Tables \ref{Table: RR100} and \ref{Table: RR500} report the performance of each configuration under 100 and 500 improvement iterations, respectively. Notably, “Time” refers to the total execution time of the improver alone and excludes the time required for initial solution construction. 
The experimental results show that increasing $k$ reduces the optimality gap but increases the execution time. Expanding the regional reconstruction scope effectively improves reconstruction quality. We select $k=60$ as a trade-off between execution time and optimality gap. 

\begin{table}[t]
\small
\centering
\caption{Effect of candidate set scale for the constructive model.}
\setlength{\tabcolsep}{0.8mm}
\begin{tabular}{l|cc|cc|cc}
\toprule
\multirow{2}{*}{Scale} 
& \multicolumn{2}{c|}{TSP1K} & \multicolumn{2}{c|}{TSP5K} & \multicolumn{2}{c}{TSP10K}\\
&Gap&Time &Gap&Time &Gap&Time\\
\midrule

$k_d=50$ &3.64\% &4s  &5.90\% &19s  &10.39\% &36s \\

$k_d=100$ &3.03\% &7s  &4.14\% &19s  &6.19\% &36s \\

$k_d=200$ &2.67\% &17s  &3.28\% &21s  &4.66\% &43s \\

$k_d=300$ &2.91\% &29s  &2.87\% &30s  &4.34\% &59s \\

$k_d=400$ &2.93\% &42s  &3.49\% &42s  &4.21\% &1.43m \\

$k_d=500$ &3.01\% &53s  &3.78\% &54s  &4.47\% &1.85m \\

$k_d=1000$ &2.48\% &1.62m  &3.81\% &2.57m  &4.76\% &5.45m \\
\bottomrule
\end{tabular}
\label{Table: KNN}
\end{table}

\textbf{Effect of Candidate Set Size on the Constructive Model}
To investigate the impact of candidate set scale, we conducted this experiment with the maximum candidate set size $k_d$ set to 50, 100, 200, 300, 400, 500, 1000. Under these settings, the performance of the constructive model on uniform TSP1K, TSP5K, and TSP10K is shown in Table \ref{Table: KNN}. 
Experimental results show that increasing $k_d$ reduces the optimality gap but increases computation time. However, once 
$k_d$ exceeds approximately 200, the gap stops decreasing consistently and begins to fluctuate. We attribute this behavior to information saturation: by that point, the model already has nearly all useful candidate information, and adding more may introduce noise. Hence, we choose $k_d=200$ as a practical compromise. 

\begin{table}[t]
\small
\centering
\caption{Performances on larger-scale TSP instances.}

\begin{tabular}{c|cc|cc}
\toprule
\multirow{2}{*}{Method} 
& \multicolumn{2}{c|}{TSP50K} & \multicolumn{2}{c}{TSP100K}\\
&Gap&Time &Gap&Time\\
\midrule





Ours greedy &7.06\% &3.61m  &10.43\% &7.6m \\

Ours Rec100 &4.32\% &7.03m &5.12\% &12.66m \\

Ours Rec500 &3.96\% &20.76m &4.74\% &33.13m \\
\bottomrule
\end{tabular}
\label{Table: TSP-5W}
\end{table}

\subsection{Performance on Larger Scale Instances} 
Table \ref{Table: TSP-5W} presents the evaluation results of LocalEscaper on TSP50K and TSP100K. 
To adapt the constructive model to these scales, we continue training it on a uniform training set of 100 instances, each containing 50,000 nodes, and employ a lightweight configuration during inference. More details can be found in the Appendix \ref{Appendix-Larger-Scale}. The result demonstrates the feasibility of our method for tackling large-scale problems. 

\section{Conclusion}
In this work, we propose LocalEscaper, a weakly-supervised framework with regional reconstruction for scalable neural TSP solvers.
Our method leverages RL-based heuristics to refine solutions and iteratively enhance low-quality training labels, thereby reducing reliance on expert-annotated data. 
For solution improvement, we introduce a regional reconstruction strategy that enhances subsequence reconstruction by improving node reordering and escaping local optima. 
The LocalEscaper framework delivers notable performance among neural combinatorial optimization solvers, outperforming previous methods in solution accuracy, inference speed, and generalization to large-scale TSP instances. 
Despite its strengths, LocalEscaper faces the limitation: its regional reconstruction mechanism may still become trapped in local optima. We plan to explore solutions to these issues in future work. In addition, our framework is potentially extensible to other combinatorial optimization problems, which we plan to explore in future work.

\bibliography{References}
\bibliographystyle{icml2025}

\newpage
~
\newpage

\appendix
\setcounter{equation}{0}
\setcounter{table}{0}
\renewcommand{\theequation}{A\arabic{equation}}
\renewcommand{\thetable}{A\arabic{table}}
\renewcommand{\thealgorithm}{A\arabic{algorithm}}

\renewcommand{\algorithmicrequire}{\textbf{Input:}}
\renewcommand{\algorithmicensure}{\textbf{Output:}}

\section{Implementation Details for Subsequence Reconstruction}
\label{Appendix-Subsequence Reconstruction}
In this section, we introduce the implementation details for the subsequence reconstruction approach. 

\subsection{Preprocessing}
We randomly decompose a TSP solution of length $n$ into $\lfloor \frac{n}{m} \rfloor$ disjoint subsequences of length $m$, each of which is then reconstructed. 

For a given subsequence $\boldsymbol{\tau}^{\prime} = (\tau_1^{\prime}, \tau_2^{\prime}, \dots, \tau_{m}^{\prime})$, let the vector $\boldsymbol{x}$ denote the $x$-coordinates of all its nodes, and the vector $\boldsymbol{y}$ denote their $y$-coordinates. 
To improve the homogeneity of the input, we apply a coordinate transformation to the $x$ and $y$ coordinates of the nodes:  
\begin{equation}
\begin{aligned}
\boldsymbol{x} \leftarrow \frac{\boldsymbol{x}-\bar{x}}{\sigma},~\boldsymbol{y}\leftarrow\frac{\boldsymbol{y}-\bar{y}}{\sigma} 
\label{eq:SR Preprocessing1}
\end{aligned}
\end{equation}
where $\bar{x}=\frac{1}{m}\sum_{i=1}^{m} x_i$ and $\bar{y}=\frac{1}{m}\sum_{i=1}^{m} y_i$ are the mean of $\boldsymbol{x}$ and $\boldsymbol{y}$, and 
\begin{equation}
\begin{aligned}
\sigma = \max\bigl(\max_{1\le i\le m}|x_i|,\;\max_{1\le i\le m}|y_i|\bigr),
\label{eq:SR Preprocessing2}
\end{aligned}
\end{equation}
is the maximum absolute coordinate value among all nodes. 

\subsection{The Markov Decision Process}
We model the subsequence reconstruction task as a Markov Decision Process, which consists of states, actions, transition, and rewards. 

\textit{States}: The state is characterized as the original subsequence $\boldsymbol{\tau}^{\prime}$ and current new subsequence $\boldsymbol{\tau}^{\prime\prime}$. Initially, $\boldsymbol{\tau}^{\prime\prime}$ is an empty list. 

\textit{Actions}: An action is the selection of the next visited node $\tau_\text{act}$. The first visited node is fixed as $\tau_{1}^{\prime}$, and the last visited node is fixed as $\tau_{m}^{\prime}$. 

\textit{Transition}: After performing an action, the node $\tau_\text{act}$ is appended to $\boldsymbol{\tau}^{\prime\prime}$, and removed from $\boldsymbol{\tau}^{\prime}$. When visiting the last node $\tau_{m}^{\prime}$, the reconstruction is completed. 

\textit{Rewards}: The total reward is equal to the total length of the new tour $\boldsymbol{\tau}^{\prime\prime}$, which can be calculated by Equation (2) in the main text. 

\subsection{Transformer Network}
\textbf{Encoder}: 
The input to the network is feature $\mathbf{f} \in \mathbb{R}^{m \times 2}$, the vector of coordinates of all nodes in the subsequence. We use a linear projection to transform each subsequence feature $\{\mathbf{f}_i\}_{i=1}^{m}$ into its embeddings $\{\mathbf{h}_i^{(0)}\}_{i=1}^{m}$, where $\mathbf{h}_i^{(0)}=\mathbf{f}_i W^{(0)}+\mathbf{b}^{(0)}$, with $W^{(0)} \in \mathbb{R}^{2 \times d}$ and $\mathbf{b}^{(0)} \in \mathbb{R}^d$ are learnable matrices, and $d$ is the embedding dimension. Then, we employ a multi-head attention layer to map the embedding matrix $H^{(0)}=\{\mathbf{h}_i^{(0)}\}_{i=1}^{m}$ into $H^{(1)}\in \mathbb{R}^{m \times d}$ via $H^{(1)}=\text{AttentionLayer}(H^{(0)})$, where AttentionLayer$(\cdot)$ denotes the attention layer. 

The attention layer contains two modules: the multi-head attention (MHA) module and feed-forward (FF) module \cite{kool2018attention}. 
Let the input and output be denoted by $H^{(l-1)} \in \mathbb{R}^{N \times d}$ and $H^{(l)}\in \mathbb{R}^{N \times d}$ (In this scenario, $N=m$), respectively.
The forward process of the attention layer is given by: 
\begin{equation}
\begin{aligned}
&\tilde{H}^{(l-1)}=\text{softmax}\left(\frac{
H^{(l-1)}W_q(H^{(l-1)}W_k)^{\text{T}}
}
{
\sqrt{d}
}\right) H^{(l-1)}W_v,
\\
&\hat{H}^{(l)}=H^{(l-1)}+\tilde{H}^{(l-1)},
\\
&H^{(l)}=\hat{H}^{(l)}+\text{FF}(\hat{H}^{(l)}),
\label{eq:SR Attention}
\end{aligned}
\end{equation}
where $W_q$, $W_k$, $W_v \in \mathbb{R}^{d \times d}$ are learnable matrices. 

The embeddings of the first and last nodes $\tau_1^{\prime}$ and $\tau_{m}^{\prime}$ in $H^{(1)}$ are denoted as $\mathbf{h}_1^{(1)}$ and $\mathbf{h}_m^{(1)}$. We use two multi-layer perceptrons (MLP) to encode $\mathbf{h}_1^{(1)}$ and $\mathbf{h}_m^{(1)}$ separately, then concatenate them and the embeddings of other nodes, and feed the result into $l_s$ multi-head attention layers. The calculation process can be described as: 
\begin{equation}
\begin{aligned}
Z&=[\text{MLP}(\mathbf{h}_1^{(1)}),
\{\mathbf{h}_i^{(1)}\}_{i=2}^{m-1},
\text{MLP}(\mathbf{h}_m^{(1)})]
\\
H^{(2)}&=\text{AttentionLayer}(Z),
\\
&...
\\
H^{(l_s+1)}&=\text{AttentionLayer}(H^{(l_s+1)}),
\label{eq:RR Attention}
\end{aligned}
\end{equation}
where $[.,.]$ denotes the vertical concatenation operator, and MLP$(\cdot)$ denotes the multi-layer perceptron module. 
Then, we use two separate multi-layer perceptrons to transform $H^{(l_S+1)}$ into two separate intermediate embedding matrices $H^{\text{A}} \in \mathbb{R}^{m \times d}$ and $H^{\text{B}} \in \mathbb{R}^{m \times d}$, and calculate the heatmap $\boldsymbol{\mathcal{H}} \in \mathbb{R}^{m \times m}$, as shown in the following equations: 
\begin{equation}
\begin{aligned}
H^{\text{A}}&=\text{MLP}(H^{(l_s+1)}),
\\
H^{\text{B}}&=\text{MLP}(H^{(l_s+1)}),
\\
\boldsymbol{\mathcal{H}}&=C \cdot \text{tanh}\left(\frac{H^{\text{A}}(H^{\text{B}})^{T}}{\sqrt{d}}\right),
\label{eq:RR heatmap}
\end{aligned}
\end{equation}
where, MLP$(\cdot)$ denotes the multi-layer perceptron module, and $C$ is a constant. The heatmap $\boldsymbol{\mathcal{H}} \in \mathbb{R}^{m \times m}$ serves to score the connections between nodes in subsequence $\boldsymbol{\tau}^{\prime} = (\tau_1^{\prime}, \tau_2^{\prime}, \dots, \tau_{m}^{\prime})$. 

\textbf{Decoder}: 
In the decoder, we calculate the selection probability of each action in the current state based on the heatmap $\boldsymbol{\mathcal{H}}$. 
At step $t$, if the last node in the current new tour $\boldsymbol{\tau}^{\prime\prime}$ is $\tau_{i}^{\prime}$, then the action scores of $\tau_{j}^{\prime}$ is calculated as follows: 
\begin{equation}
u_{i,j}=\left\{
\begin{aligned}
\boldsymbol{\mathcal{H}}_{i,j},~~~~&\tau_{j}^{\prime} \notin \boldsymbol{\tau}^{\prime\prime}~\text{and}~\tau_{j}^{\prime}\neq\tau_{m}^{\prime}
\\
-\infty,~~~~~~&\text{otherwise}
\label{eq:RR mask}
\end{aligned}
\right.
~,
\end{equation}
where $u_{i,j}$ denotes the scores for transitioning from $\tau_{i}^{\prime}$ to $\tau_{j}^{\prime}$. 
We calculate the scores $\{u_{i,j}\}_{j=1}^{m}$ for current node $\tau_{i}^{\prime}$. 
Finally, a softmax function is applied to $\{u_{i,j}\}_{j=1}^{m}$ to produce the selection probability of each action. 

We sample actions based on probabilities to complete the subsequence reconstruction following the Markov decision process introduced in the previous subsection. 
For a reconstructed subsequence $\boldsymbol{\tau}^{\prime\prime} = (\tau_1^{\prime\prime}, \tau_2^{\prime\prime}, \dots, \tau_{m}^{\prime\prime})$, if its length $L_{\text{sub}}(\boldsymbol{\tau}^{\prime\prime})$ is shorter than the original subsequence length $L_{\text{sub}}(\boldsymbol{\tau}^{\prime})$, it replaces the original subsequence.

We follow POMO’s multiple‐trajectory approach \cite{kwon2020pomo}. 
The model is trained by sampling multiple trajectory lists $\boldsymbol{\mathcal{T}}^{\prime\prime}=\{\boldsymbol{\tau}^{\prime\prime}_{(1)}, \boldsymbol{\tau}^{\prime\prime}_{(2)}, \dots, \boldsymbol{\tau}^{\prime\prime}_{(N)} \}$, where each $\boldsymbol{\tau}^{\prime\prime}_{(i)}$ represents a possible reconstructed subsequence generated by the heatmap. 

The gradient of the policy network is approximated as: 
\begin{equation}
\begin{aligned}
\triangledown_{\boldsymbol{\psi}}J(\boldsymbol{\psi}) &\approx \frac{1}{N}\sum\limits_{i=1}^{N} \frac{L_{\text{sub}}(\boldsymbol{\tau}^{\prime\prime}_{(i)}) - \mu(\boldsymbol{\mathcal{T}}^{\prime\prime})}{\delta(\boldsymbol{\mathcal{T}}^{\prime\prime})} \triangledown_{\boldsymbol{\psi}} \text{log} ~ p_{\boldsymbol{\psi}}(\boldsymbol{\tau}^{\prime\prime}_{(i)}| \boldsymbol{\tau}^{\prime}),
\label{eq:Regional Reconstruction Gradient}
\end{aligned}
\end{equation}
Here, $p_{\boldsymbol{\psi}}(\boldsymbol{\tau}^{\prime\prime}{(i)},|,\boldsymbol{\tau}^{\prime})$ denotes the probability of model $\boldsymbol{\psi}$ constructing the solution $\boldsymbol{\tau}^{\prime\prime}{(i)}$ given the subsequence reconstruction problem $\boldsymbol{\tau}^{\prime}$.
The symbols $\mu(\boldsymbol{\mathcal{T}}^{\prime\prime})$ and $\delta(\boldsymbol{\mathcal{T}}^{\prime\prime})$ represent the mean and standard deviation, respectively, of the rewards (the list of lengths) $\{L_{\text{sub}}(\boldsymbol{\tau}^{\prime\prime}_{(1)}), \dots, L_{\text{sub}}(\boldsymbol{\tau}^{\prime\prime}_{(N)})\}$ associated with the trajectories $\boldsymbol{\mathcal{T}}^{\prime\prime}$. We set $N = 128$.

At the inference stage, we select the best trajectory.

\section{Implementation Details for Regional Reconstruction}
\label{Appendix-Regional Reconstruction}
In this section, we introduce the implementation details for the regional reconstruction approach. 

\subsection{Problem Setup}
Given an instance graph $\mathcal{G}$ and its current solution $\boldsymbol{\tau} = (\tau_1, \tau_2, \dots, \tau_n)$, the corresponding set of edges $\mathcal{E}_{\boldsymbol{\tau}}=\{e_{\tau_{i},\tau_{i+1}} | 1\leq i \leq n-1\} \cup \{e_{\tau_{n},\tau_{1}}\}$ represents the set of connections in solution $\boldsymbol{\tau}$. 
We uniformly random sample a coordinate $c = (x, y)$ in the coordinate space. The node set $\mathcal{V}_{c}$ represents the $k$-nearest neighbors of $c$. According to the current solution $\boldsymbol{\tau}$, the edges in $\mathcal{E}_{\boldsymbol{\tau}}$ that have nodes from $\mathcal{V}_{c}$ as their predecessors are defined as $\mathcal{E}_{\boldsymbol{\tau}}^{\prime}=\{e_{\tau_{i},\tau_{i+1}} | \tau_{i} \in \mathcal{V}_{c}\}$. Specifically, define $\tau_{n+1}=\tau_{1}$ to denote the successor of $\tau_{n}$. 

We remove $\mathcal{E}_{\boldsymbol{\tau}}^{\prime}$ from $\mathcal{E}_{\boldsymbol{\tau}}$, and the remaining edges $\mathcal{E}_{\boldsymbol{\tau}} \setminus \mathcal{E}_{\boldsymbol{\tau}}^{\prime}$ can compose $k$ subsequences 
$\mathcal{S}=\{ \boldsymbol{\tau}_{1}^{\prime}, \boldsymbol{\tau}_{2}^{\prime}, \dots, \boldsymbol{\tau}_{k}^{\prime} \}$
, where each subsequence is independent of the direction of traversal. Then, We permute subsequences in $\mathcal{S}$ into a list $\mathcal{X}=(\boldsymbol{\eta}_{1}, \boldsymbol{\eta}_{2}, \dots, \boldsymbol{\eta}_{k})$, allowing any subsequence to be reversed during permutation. For instance, for a subsequence $\boldsymbol{\tau}^{\prime} = (\tau_1^{\prime}, \tau_2^{\prime}, \dots, \tau_{m}^{\prime})$, its reverse is denoted as $\neg \boldsymbol{\tau}^{\prime} = (\tau_{m}^{\prime}, \tau_{m-1}^{\prime}, \dots, \tau_{1}^{\prime})$. Whether reversed or not, the path from $\tau_1^{\prime}$ to 
$\tau_m^{\prime}$ remains fixed, as does the path from $\tau_m^{\prime}$ to 
$\tau_1^{\prime}$, and the lengths of these two paths are equal. 
Therefore, during permutation, we only need to focus on the information of the first and last nodes of the subsequence, and the information of intermediate nodes can be ignored. 

\subsection{Transformer Network}
\textbf{Encoder}: In the MDP of regional reconstruction task, the initial set of available subsequences is $\mathcal{C}_{\text{init}}=\{\boldsymbol{\tau}_{1}^{\prime}, \neg\boldsymbol{\tau}_{1}^{\prime}, \boldsymbol{\tau}_{2}^{\prime}, \neg\boldsymbol{\tau}_{2}^{\prime},\dots,\boldsymbol{\tau}_{k}^{\prime}, \neg\boldsymbol{\tau}_{k}^{\prime}\}$. We use a 4D vector to represent the features of a subsequence. For a subsequence $\boldsymbol{\tau}^{\prime} = (\tau_1^{\prime}, \tau_2^{\prime}, \dots, \tau_{m}^{\prime})$
, its features is $(x_{\tau_1^{\prime}}, y_{\tau_1^{\prime}}, x_{\tau_m^{\prime}}, y_{\tau_m^{\prime}})$, which represent the $x$ and $y$ coordinates of the first node $\tau_1^{\prime}$ and last nodes $\tau_m^{\prime}$. 
We feed the features of all subsequences in $\mathcal{C}_{\text{init}}$ into the encoder as tokens. To improve the homogeneity of the input, we preprocess the node coordinates using Equation \eqref{eq:SR Preprocessing1}. 

We use a linear projection to transform each subsequence feature $\{\mathbf{f}_i\}_{i=1}^{2k}$ into its embeddings $\{\mathbf{h}_i^{(0)}\}_{i=1}^{2k}$, where $\mathbf{h}_i^{(0)}=\mathbf{f}_i W^{(0)}+\mathbf{b}^{(0)}$, with $W^{(0)} \in \mathbb{R}^{4 \times d}$ and $\mathbf{b}^{(0)} \in \mathbb{R}^d$ are learnable matrices, and $d$ is the embedding dimension. 

Similar to \cite{kwon2020pomo}, we employ $l_r$ multi-head attention layers to map the embedding matrix $H^{(0)}=\{\mathbf{h}_i^{(0)}\}_{i=1}^{2k}$ into $H^{(l_r)}=\{\mathbf{h}_i^{(l_r)}\}_{i=1}^{2k}$. 
The calculation process can be described as: 
\begin{equation}
\begin{aligned}
H^{(1)}&=\text{AttentionLayer}(H^{(0)}),
\\
&...
\\
H^{(l_r)}&=\text{AttentionLayer}(H^{(l_r-1)}),
\label{eq:RR Attention}
\end{aligned}
\end{equation}
where $H^{(i)} \in \mathbb{R}^{2k \times d}$ is the output of the $i$-th attention layer. 
Then, we use two separate multi-layer perceptrons to transform $H^{(l_r)}$ into two separate intermediate embedding matrices $H^{\text{A}} \in \mathbb{R}^{2k \times d}$ and $H^{\text{B}} \in \mathbb{R}^{2k \times d}$, and calculate the heatmap $\boldsymbol{\mathcal{H}} \in \mathbb{R}^{2k \times 2k}$, as shown in the following equations: 
\begin{equation}
\begin{aligned}
H^{\text{A}}&=\text{MLP}(H^{(l_r)}),
\\
H^{\text{B}}&=\text{MLP}(H^{(l_r)}),
\\
\boldsymbol{\mathcal{H}}&=C \cdot \text{tanh}\left(\frac{H^{\text{A}}(H^{\text{B}})^\text{T}}{\sqrt{d}}\right),
\label{eq:RR heatmap}
\end{aligned}
\end{equation}
where, MLP$(\cdot)$ denotes the multi-layer perceptron module, and $C$ is a constant. The heatmap $\boldsymbol{\mathcal{H}} \in \mathbb{R}^{2k \times 2k}$ serves to score the connections between subsequences in $\mathcal{C}_{\text{init}}$, which the decoder then uses to calculate the selection probabilities of those connections. 

\textbf{Decoder}: 
In the decoder, we calculate the selection probability of each action in the current state based on the heatmap $\boldsymbol{\mathcal{H}}$. 
For simplicity, we write $\mathcal{C}_{\text{init}}$ alternatively as $\mathcal{C}_{\text{init}}=\{\hat{\boldsymbol{\tau}}_{1}, \hat{\boldsymbol{\tau}}_{2},\dots, \hat{\boldsymbol{\tau}}_{2k}\}$, where for each $i \in \{1,2,\dots,k\}$, $\hat{\boldsymbol{\tau}}_{2i-1}=\boldsymbol{\tau}_{i}^{\prime}$ and $\hat{\boldsymbol{\tau}}_{2i}=\neg\boldsymbol{\tau}_{i}^{\prime}$. 

If the last subsequence in the current list $\mathcal{X}$ is $\hat{\boldsymbol{\tau}}_{i}$, then the action scores of $\hat{\boldsymbol{\tau}}_{j}$ is calculated as follows: 
\begin{equation}
u_{i,j}=\left\{
\begin{aligned}
\boldsymbol{\mathcal{H}}_{i,j},~~~~&\hat{\boldsymbol{\tau}}_{j} \notin \mathcal{X}
\\
-\infty,~~~~~~&\text{otherwise}
\label{eq:RR mask}
\end{aligned}
\right.
~,
\end{equation}
where $u_{i,j}$ denotes the scores for transitioning from $\hat{\boldsymbol{\tau}}_{i}$ to $\hat{\boldsymbol{\tau}}_{j}$. 
We calculate the scores $\{u_{i,j}\}_{j=1}^{2k}$ for last subsequence $\hat{\boldsymbol{\tau}}_{i}$.  
Finally, a softmax function is applied to $\{u_{i,j}\}_{j=1}^{2k}$ to produce the selection probability of each action. 

For a reconstructed tour $\boldsymbol{\tau}^\text{new}$, if its length $L_{\text{total}}(\boldsymbol{\tau}^\text{new})$ is shorter than the original subsequence length $L_{\text{total}}(\boldsymbol{\tau})$, it replaces the original tour.

We follow POMO’s multiple‐trajectory approach \cite{kwon2020pomo}. At each step, we sample 128 trajectories and train the network using equation (4) in main text. At inference stage, we select the best trajectory. 

\section{Implementation Details for Lightweight Constructive Model}
\label{Appendix-Lightweight Model}
In this section, we introduce the implementation details for the lightweight constructive mode. 

\textbf{Encoder}: For a $n$ nodes TSP graph $\mathcal{G}(\mathcal{V}, \mathcal{E})$, we use a 2-layer perceptron to transform the input from features $\mathbf{f}\in \mathbb{R}^{n \times 2}$ into embeddings $\mathbf{h}^{(0)} \in \mathbb{R}^{n \times d}$, where 
\begin{equation}
\begin{aligned}
\mathbf{h}^{(0)}&=\text{MLP}(\mathbf{f}).
\label{eq:lightweight encoder}
\end{aligned}
\end{equation}

\textbf{Decoder}: Let $\tau_\text{last}$ denote the last visited node and $\tau_\text{first}$ denote the first visited node. We select the $k_d$ nearest unvisited nodes of $\tau_\text{last}$ as the candidate set $\mathbb{A}^\text{p}$. 
Then we feed candidates $\mathbb{A}^\text{p}$, the last visited node $\tau_\text{last}$ and the first visited node $\tau_\text{first}$ into the decoder. The decoder concatenates these embeddings and feeds them into $l_c$ attention layers as follows: 
\begin{equation}
\begin{aligned}
Z&=[\text{MLP}(\mathbf{h}_{\text{first}}^{(0)}),
\{\mathbf{h}_{\mathbb{A}^\text{p}}^{(1)}\}_{i=1}^{k_d},
\text{MLP}(\mathbf{h}_{\text{last}}^{(1)})]
\\
H^{(1)}&=\text{AttentionLayer}(Z),
\\
&...
\\
H^{(l_c)}&=\text{AttentionLayer}(H^{(l_c-1)}),
\label{eq:lightweight decoder}
\end{aligned}
\end{equation}
where $H^{(l_c)} \in \mathbb{R}^{(|\mathbb{A}^\text{p}|+2) \times d}$ is the output of the $i$-th attention layer. 

Then, $H^{(l_c)}$ is map to a score vector $H^{(l_c+1)} \in \mathbb{R}^{(|\mathbb{A}^\text{p}|+2) \times 1}$ via a linear projection, as follows:  
\begin{equation}
\begin{aligned}
H^{(l_c+1)}=H^{(l_c)} W+\mathbf{b}
\label{eq:lightweight decoder}
\end{aligned}
\end{equation}
where $W \in \mathbb{R}^{d \times 1}$ and $\mathbf{b} \in \mathbb{R}^1$ are learnable parameters. Finally, we apply a softmax function to $H^{(l_c+1)}$ to produce the selection probability of candidates $\mathbb{A}^\text{p}$. Notably, the first visited node is fixed at the depot and does not require the model to compute selection probabilities. 

The model discards the information of distant nodes, concentrating computation on high-value, nearby nodes. This focus substantially reduces the computational complexity of the attention mechanism on large-scale problems, lowering it from $O(n^2)$ to $O(|\mathbb{A}^\text{p}|^2)$. 

During training, if the number of unvisited nodes does not exceed 1,000, we do not prune the candidate set; instead, we include all unvisited nodes as candidates. This approach fully leverages large-neighborhood information, allowing the model to predict optimal selections over a wider range, thereby enforcing stricter training and enhancing model performance.

\begin{table}[ht]
\centering
\caption{Setting of the constructive model.}
\setlength{\tabcolsep}{0.8mm}
\begin{tabular}{l|c}
\toprule
Parameter & Value \\
\midrule
Training Epoch &300 \\
Training Batch Size &64 \\
Learning Rate &1e-4 \\
Decay per Training Epoch &0.98 \\
Number of Encoder layers &1 \\
Number of Dncoder layers $l_c$ &6 \\
Candidate Size $k_d$ &200 \\
\bottomrule
\end{tabular}
\label{Tab_Model_setting1}
\end{table}

\begin{table}[ht]
\centering
\caption{Setting of the improver (Consisting of the subsequence and regional reconstruction models).}
\setlength{\tabcolsep}{0.8mm}
\begin{tabular}{l|c}
\toprule
Parameter & Value \\
\midrule
Improvement Iteration for Training &1200\\
Training Epochs per Iteration &20 \\
Training Batch Size &128 \\
Learning Rate &1e-4 \\
Decay per Training Epoch &1 \\
Number of Encoder layers $l_s$(or $l_r$) &6 \\
Number of Dncoder layers &1 \\
Candidate Size $k_d$ &200 \\
Subsequence Reconstruction Length $m$ &100 \\
Regional Reconstruction Scale $k$ &60 \\
\bottomrule
\end{tabular}
\label{Tab_Model_setting2}
\end{table}

\begin{algorithm}[ht]
    \small
	\caption{Training} 
	\label{alg: Training} 
	\begin{algorithmic}
		\REQUIRE Dataset $\mathcal{D}$ containing $D$ instances (denoted as $\boldsymbol{\mathcal{G}}$); Maximum number of improvement iterations: $T$; Training epochs per iteration for the improver: $I_\text{train}$; Maximum training epoch for the constructive model: $C_\text{train}$
		\ENSURE The parameters of the constructive model $\boldsymbol{\theta}$, the subsequence reconstruction model $\boldsymbol{\psi}$ and the regional reconstruction model $\boldsymbol{\phi}$
		\STATE Initialize the parameters of all model
        \STATE Generate the initial labels $\boldsymbol{\mathcal{T}}^0$: 
        \\$\{\boldsymbol{\tau}_1,\dots,\boldsymbol{\tau}_D\} \gets \text{RandomInsertion}(\boldsymbol{\mathcal{G}})$ 
        \STATE Initialize dataset $\mathcal{D}^0 \gets \{\boldsymbol{\mathcal{G}},\boldsymbol{\mathcal{T}}^0\}$
        \STATE Execute the following two processes in parallel:
        \STATE Processes 1:
		\FOR{$t=1:T$}
        \STATE $\boldsymbol{\mathcal{T}}^t \gets \text{SubsequenceReconstruction}(\boldsymbol{\mathcal{G}},\boldsymbol{\mathcal{T}}^{t-1},\boldsymbol{\psi})$
        \STATE $\boldsymbol{\mathcal{T}}^t \gets \text{2opt}(\boldsymbol{\mathcal{G}},\boldsymbol{\mathcal{T}}^{t})$
        \STATE $\boldsymbol{\mathcal{T}}^t \gets \text{RegionalReconstruction}(\boldsymbol{\mathcal{G}},\boldsymbol{\mathcal{T}}^{t},\boldsymbol{\phi})$
        \STATE Update dataset $\mathcal{D}^t \gets \{\boldsymbol{\mathcal{G}},\boldsymbol{\mathcal{T}}^t\}$.

        \FOR{$i=1:I_\text{train}$}
        \STATE $\boldsymbol{\psi} \gets \text{SubsequenceModelTraining}(\boldsymbol{\mathcal{G}},\boldsymbol{\mathcal{T}}^{t},\boldsymbol{\psi})$
        \STATE $\boldsymbol{\phi} \gets \text{RegionalModelTraining}(\boldsymbol{\mathcal{G}},\boldsymbol{\mathcal{T}}^{t},\boldsymbol{\phi})$
        \ENDFOR
        \ENDFOR

        \STATE Processes 2:
        \FOR{$i=1:C_\text{train}$}
        \STATE Load the latest dataset $\mathcal{D}^*(\boldsymbol{\mathcal{G}},\boldsymbol{\mathcal{T}}^*)$
        \STATE $\boldsymbol{\theta} \gets \text{ConstructiveModelTraining}(\boldsymbol{\mathcal{G}},\boldsymbol{\mathcal{T}}^*,\boldsymbol{\theta})$
        \ENDFOR 
	\end{algorithmic} 
\end{algorithm}

\section{Details of Model Settings}
\label{Appendix-Training setting}Our framework consists of three models: a constructive model, a subsequence reconstruction model, and a regional reconstruction model. We summarize the settings of the constructive model in Table~\ref{Tab_Model_setting1}, and those of the improver—which consists of the subsequence and regional reconstruction models—in Table~\ref{Tab_Model_setting2}.

When training the constructive model, the label length $|\boldsymbol{\tau}^{\prime}|$ is sampled from the range $[50, 100]$ for each batch. After each epoch, the upper bound of this range is increased by 5, until it reaches 1000 nodes.

During the training of the improver, we alternate between dataset improvement and model training. Specifically, after one improvement iteration over the entire dataset, both the subsequence and regional reconstruction models are trained for 20 epochs. In each epoch, we split all instances into multiple subsequences as described in Section \ref{Appendix-Subsequence Reconstruction}, which are then used to train the subsequence reconstruction model. For the regional reconstruction model, we construct one regional reconstruction task for each instance following the procedure in Section \ref{Appendix-Regional Reconstruction} and use them for training. 

Both the constructive model and the improver are trained on the same dataset. 
We adopt the standard data generation process from the previous work~\cite{kool2018attention}. 
The training set consists of 8192 instances, each with 1000 nodes sampled from a uniform distribution.

\begin{algorithm}[t]
    \small
	\caption{Inference} 
	\label{alg: Inference} 
	\begin{algorithmic} 
    \REQUIRE Instance $\mathcal{G}$; The parameters of the constructive model $\boldsymbol{\theta}$, the subsequence reconstruction model $\boldsymbol{\psi}$ and the regional reconstruction model $\boldsymbol{\phi}$; Maximum number of improvement iterations $T$
    \ENSURE The best solution $\boldsymbol{\tau}^T$
    \STATE $\boldsymbol{\tau}^0 \gets \text{ConstructSolutionByModel}(\mathcal{G},\boldsymbol{\theta})$
    \FOR{$t=1:T$}
    \STATE $\boldsymbol{\tau}^t \gets \text{SubsequenceReconstruction}(\mathcal{G},\boldsymbol{\tau}^{t-1},\boldsymbol{\psi})$
    \STATE $\boldsymbol{\tau}^t \gets \text{2opt}(\mathcal{G},\boldsymbol{\tau}^{t})$
    \STATE $\boldsymbol{\tau}^t \gets \text{RegionalReconstruction}(\mathcal{G},\boldsymbol{\tau}^{t},\boldsymbol{\phi})$
    \ENDFOR 
    \end{algorithmic} 
\end{algorithm}

\section{Pseudocode}
\label{Appendix-Pseudocode}
This section provides the pseudocode of LocalEscaper for better understanding. Algorithm \ref{alg: Training} is for model training. Algorithm \ref{alg: Inference} is for inference.

\section{Results on TSPLIB}
\label{Appendix-TSPLIB}

Table \ref{Table: TSPLIB} presents the experimental results of each algorithm on the real-world TSPLIB dataset, with instances categorized into groups based on size.The "Obj." represents the average length of the instances calculated by the method, "Gap" is the average of the Gap values individually calculated for each instance, and "Time" denotes the total computation time. Table \ref{Table: TSPLib1} and Table \ref{Table: TSPLib2} provide detailed results for each instance. Some baselines, using code provided by the original papers, failed to run on certain instances; these results are marked as "ERR." Our method achieved the best results, recording the lowest gap across all three groups while demonstrating exceptionally fast solving speeds.

\section{Experimental Settings for Larger-Scale Instances}
\label{Appendix-Larger-Scale} 
We prepare two additional large-scale test sets: TSP50K and TSP100K. These datasets follow the same generation standards as prior work \cite{kool2018attention} and use LKH to compute ground-truth labels. We also create a training set of 96 instances, each with 50,000 nodes, to further adapt the constructive model to large-scale data. We continue training it under our proposed framework, starting from low-quality labels with a learning rate of 1e-5 for 500 epochs. The subsequence and regional reconstruction models are not retrained. To accommodate runtime and memory constraints, we omit the 2-opt step in the improver. For TSP100K inference, we set the subsequence reconstruction length $m$ to 50. 

\begin{table*}[ht]
\centering
\small
\caption{Detailed results on TSPLIB benchmark instances (Part I).}
\setlength{\tabcolsep}{0.5mm}
\begin{tabular}{c c c c c c c c c c c c c c c c c c c c c c c c}
\toprule
Instance & Optimal & \multicolumn{3}{c}{Pointerformer} & \multicolumn{3}{c}{GLOP} & \multicolumn{3}{c}{Att-GCN} & \multicolumn{3}{c}{DIFUSCO} & \multicolumn{3}{c}{LEHD RRC100} & \multicolumn{3}{c}{SIT PRC 100} & \multicolumn{3}{c}{Ours Rec 500} \\
Name     &         & Gap & Time(s) & & Gap & Time(s) & & Gap & Time(s) & & Gap & Time(s) & & Gap & Time(s) & & Gap & Time(s) & & Gap & Time(s) \\
\midrule
eil51    & 426    & 0.47 & 0.71 & & 0.23 & 1.94  & & 0.23 & 5.15 & & 0.23 & 7.10 & & 0.23 & 16.72 & & 0.47 & 20.58 & & 0.47 & 26.56 \\
berlin52 & 7542   & 0.00 & 0.77 & & 0.00 & 1.34  & & 0.00 & 5.25 & & 0.00 & 7.12 & & 0.01 & 28.14 & & 0.00 & 18.93 & & 0.00 & 26.80 \\
st70     & 675    & 0.00 & 0.72 & & 0.74 & 1.31  & & 0.15 & 7.05 & & 0.15 & 9.72 & & 0.59 & 35.22 & & 0.15 & 25.83 & & 0.15 & 31.63 \\
eil76    & 538    & 0.00 & 0.74 & & 1.86 & 1.33  & & 0.00 & 7.66 & & 0.56 & 9.61 & & 1.30 & 39.01 & & 0.00 & 27.54 & & 0.00 & 32.59 \\
pr76     & 108159 & 0.89 & 1.15 & & 0.65 & 1.34  & & 0.00 & 7.65 & & 0.00 & 9.60 & & 0.00 & 47.96 & & 0.14 & 25.26 & & 0.14 & 32.55 \\
rat99    & 1211   & 3.88 & 1.15 & & 1.57 & 1.38  & & 0.00 & 9.97 & & 0.00 & 11.98 & & 0.17 & 43.29 & & 0.17 & 37.11 & & 0.17 & 36.21 \\
kroC100  & 20749  & 1.01 & 1.03 & & 0.00 & 2.97  & & 0.00 & 10.06 & & 0.10 & 12.02 & & 0.00 & 41.07 & & 0.00 & 36.38 & & 0.00 & 36.40 \\
kroA100  & 21282  & 0.80 & 0.87 & & 0.00 & 2.87  & & 0.00 & 10.06 & & 0.00 & 12.04 & & 0.00 & 44.72 & & 0.00 & 36.02 & & 0.00 & 36.27 \\
kroB100  & 22141  & 1.98 & 0.79 & & 0.27 & 2.90  & & 0.00 & 10.06 & & 0.00 & 11.95 & & 0.00 & 40.87 & & 0.32 & 34.72 & & 0.32 & 36.46 \\
kroE100  & 22068  & 1.33  & 0.81  && 0.29  & 2.91  && 0.00     & 10.07 && 0.17  & 12.01 && 0.05  & 39.91 && 0.42  & 38.31 && 0.22  & 36.56 \\
rd100    & 7910   & 0.00     & 0.81  && 0.01  & 2.85  && 0.00    & 10.06 && 0.04  & 12.00    && 0.00    & 33.63 && 0.00     & 34.05 && 0.00     & 36.59 \\
kroD100  & 21294  & 2.75  & 0.88  && 0.42  & 2.83  && 0.00     & 10.06 && 0.00     & 12.02 && 0.00     & 33.69 && 0.00     & 42.25 && 0.00     & 36.36 \\
eil101   & 629    & 0.16  & 1.06  && 0.48  & 2.85  && 0.00     & 10.20 && 0.64  & 12.07 && 0.00     & 33.71 && 1.75  & 39.54 && 0.16  & 36.40 \\
lin105   & 14379  & 5.10  & 0.79  && 0.00     & 2.84  && 0.00     & 10.57 && 0.00     & 13.50 && 0.00     & 29.27 && 0.00     & 38.52 && 0.00     & 36.19 \\
pr107    & 44303  & 1.17  & 0.87  && 0.00     & 2.85  && 0.00    & 10.77 && 0.30  & 12.68 && 0.30  & 36.02 && \text{ERR.} & \text{ERR.} && 0.40  & 36.32 \\
pr124    & 59030  & 0.00     & 0.94  && 0.00     & 2.92  && 0.00     & 12.48 && 0.00     & 15.61 && 0.00     & 38.54 && 0.88  & 42.94 && 0.00     & 36.17 \\
bier127  & 118282 & 10.25 & 0.83  && 2.11  & 2.91  && 0.00     & 12.78 && 0.00     & 15.94 && 0.14  & 40.36 && 0.09  & 41.52 && 0.00     & 36.32 \\
ch130    & 6110   & 0.39  & 0.89  && 0.23  & 3.00  && 0.02  & 13.08 && 0.85  & 16.33 && 0.02  & 40.82 && 0.00     & 48.33 && 0.25  & 36.57 \\
pr136    & 96772  & 0.59  & 0.92  && 0.22  & 2.92  && 0.37  & 13.69 && 0.01  & 16.88 && 0.00     & 41.17 && 0.43  & 46.22 && 0.33  & 36.01 \\
pr144    & 58537  & 1.08  & 1.14  && 0.09  & 2.96  && 0.00  & 14.50 && 0.00  & 16.55 && 0.09  & 42.13 && 0.09  & 50.98 && 0.00  & 36.15 \\
kroA150  & 26524  & 4.97  & 0.96  && 0.97  & 3.56  && 0.00  & 15.10 && 0.21  & 17.17 && 0.00  & 47.48 && 0.00  & 54.02 && 0.00  & 36.29 \\
kroB150  & 26130  & 6.85  & 1.14  && 0.05  & 3.37  && 0.01  & 15.10 && 0.05  & 17.11 && 0.01  & 44.35 && 0.01  & 53.09 && 0.06  & 36.46 \\
ch150    & 6528   & 0.81  & 0.92  && 0.32  & 3.40  && 0.23  & 15.09 && 0.32  & 17.08 && 0.00  & 45.91 && 0.00  & 48.38 && 0.40  & 36.67 \\
pr152    & 73682  & 2.27  & 0.94  && 0.45  & 3.38  && 0.27  & 15.31 && 0.00  & 17.28 && 0.82  & 49.41 && 0.76  & 53.60 && 0.00  & 36.63 \\
u159     & 42080  & 0.72  & 0.91  && 1.85  & 3.45  && 0.00  & 16.00 && 0.00  & 18.10 && 0.75  & 50.24 && \text{ERR.} & \text{ERR.} && 0.00  & 36.51 \\
rat195   & 2323   & 8.95  & 0.88  && 1.85  & 3.46  && 0.65  & 19.64 && 0.60  & 22.26 && 0.82  & 58.44 && 0.43  & 65.65 && 0.77  & 36.40 \\
d198     & 15780  & 100.41& 0.92  && 1.68  & 3.49  && 0.20  & 19.93 && 0.02  & 22.08 && 1.06  & 60.54 && 0.37  & 70.45 && 0.74  & 36.30 \\
kroB200  & 29437  & 6.16  & 1.20  && 0.12  & 3.52  && 0.00  & 20.14 && 0.04  & 23.08 && 0.00  & 59.99 && 0.00  & 75.17 && 0.00  & 36.95 \\
kroA200  & 29368  & 4.71  & 1.33  && 0.93  & 3.48  && 0.29  & 20.13 && 0.98  & 22.19 && 0.28  & 59.15 && 0.00  & 71.27 && 0.05  & 37.11 \\
ts225    & 126643 & 2.14  & 1.08  && 1.44  & 3.51  && 0.12  & 22.65 && 0.00  & 25.66 && 0.28  & 71.59 && 0.40  & 75.71 && 0.00  & 37.08 \\
tsp225   & 3916   & 25.82 & 0.89  && 1.51  & 3.52  && 0.15  & 22.65 && 0.10  & 24.77 && 0.00  & 71.66 && 1.40  & 70.74 && 1.17  & 37.07 \\
pr226    & 80369  & 3.50  & 0.93  && 0.01  & 3.48  && 0.41  & 22.75 && 0.34  & 25.81 && 1.11  & 76.81 && 0.01  & 73.99 && 0.12  & 37.01 \\
gil262   & 2378   & 2.69  & 1.12  && 1.47  & 3.55  && 0.55  & 26.38 && 0.04  & 28.51 && 1.18  & 74.82 && 0.42  & 91.87 && 0.08  & 37.11 \\
pr264    & 49135  & 14.23 & 1.41  && 0.41  & 3.54  && 0.00  & 26.57 && 0.09  & 29.19 && 5.49  & 83.48 && 0.09  & 85.65 && 0.53  & 37.27 \\
a280     & 2579   & 12.33 & 1.33  && 1.78  & 3.53  && 0.58  & 28.19 && 0.00  & 30.37 && 2.56  & 76.51 && 0.00  & 96.25 && 0.00  & 37.24 \\
pr299    & 48191  & 72.74 & 1.06  && 0.74  & 3.56  && 0.57  & 30.10 && 0.88  & 32.33 && 2.80  & 83.23 && 0.00  & 103.24&& 0.21  & 36.80 \\
lin318   & 42029  & 5.87  & 0.99  && 2.20  & 3.54  && 1.06  & 32.02 && 0.70  & 34.24 && 1.38  & 97.43 && \text{ERR.} & \text{ERR.} && 0.30  & 37.36 \\
rd400    & 15281  & 2.95  & 1.09  && 1.98  & 3.52  && 1.41  & 40.29 && 0.56  & 42.67 && 1.03  & 118.48&& 0.46  & 143.26&& 0.23  & 37.60 \\
fl417    & 11861  & 13.92 & 1.19  && 1.30  & 3.57  && 11.51 & 42.00&& 1.74  & 44.36 && 7.59  & 130.04&& 1.26  & 145.28&& 2.77  & 37.66 \\
pr439    & 107217 & 20.41 & 1.08  && 2.57  & 3.56  && 1.42  & 44.22 && 0.90  & 46.66 && 3.37  & 134.96&& 0.55  & 141.66&& 0.24  & 37.43 \\
pcb442   & 50778  & 5.17  & 1.40  && 4.09  & 3.55  && 0.67  & 44.53 && 1.01  & 47.00 && 3.10  & 141.32&& 0.03  & 166.36&& 1.03  & 37.48 \\
d493     & 35002  & 136.54& 2.32  && 2.73  & 3.56  && 1.61  & 49.68 && 1.61  & 52.26 && 9.46  & 155.77&& 0.96  & 191.43&& 1.43  & 37.56 \\
u574    & 36905   & 32.5   & 1.82  && 3.43   & 3.54  && 1.83  & 57.87  && 0.71  & 60.87  && 2.66  & 188.74  && 0.62  & 211.84  && 0.76  & 37.73  \\
rat575  & 6773    & 37.3   & 1.88  && 3.47   & 3.54  && 2.20  & 57.95  && 0.63  & 60.85  && 2.70  & 189.26  && 0.62  & 203.07  && 0.31  & 37.43  \\
p654    & 34643   & 12.55  & 2.73  && 1.58   & 3.63  && 11.62 & 65.96  && 3.22  & 69.80  && 3.28  & 191.48  && 0.18  & 223.67  && 0.12  & 37.52  \\
d657    & 48912   & 31.59  & 2.76  && 2.62   & 3.59  && 1.44  & 66.31  && 1.17  & 69.15  && 8.06  & 187.95  && 0.80  & 235.40  && 0.38  & 37.67  \\
u724    & 41910   & 20.18  & 2.83  && 3.96   & 3.66  && 2.01  & 73.11  && 0.69  & 77.09  && 3.29  & 263.70  && 0.47  & 267.80  && 0.35  & 37.83  \\
rat783  & 8806    & 37.92  & 4.01  && 4.24   & 3.57  && 2.42  & 79.15  && 1.07  & 82.14  && 3.58  & 280.43  && 0.98  & 273.07  && 1.16  & 37.68  \\
pr1002  & 259045  & 23.11  & 5.74  && 3.24   & 3.62  && 1.92  & 101.60 && 2.46  & 108.27 && 4.43  & 297.93  && 0.49  & 379.42  && 0.97  & 37.70  \\
u1060   & 224094  & 40.83  & 6.98  && 3.36   & 3.64  && 2.10  & 107.33 && 1.64  & 113.75 && 10.01 & 326.80  && 0.60  & 385.13  && 0.71  & 37.66  \\
vm1084  & 239297  & 39.82  & 7.29  && 4.91   & 3.62  && 2.53  & 109.90 && 2.70  & 116.35 && 5.42  & 344.90  && 0.80  & 387.92  && 1.20  & 37.90  \\
pcb1173 & 56892   & 17.03  & 9.50  && 6.55   & 3.70  && 2.52  & 119.36 && 2.80  & 126.10 && 7.95  & 378.68  && 0.88  & 423.98  && 1.67  & 38.15  \\
d1291   & 50801   & 33.32  & 12.60 && 7.10   & 3.78  && 3.11  & 131.16 && 3.17  & 138.78 && 13.46 & 406.63  && 1.25  & 378.63  && 1.70  & 38.31  \\
rl1304  & 252948  & 37.23  & 13.07 && 10.02  & 3.77  && 3.54  & 132.71 && 1.55  & 140.09 && 8.14  & 398.24  && 0.51  & 398.14  && 1.97  & 38.25  \\
rl1323  & 270199  & 31.05  & 9.65  && 7.96   & 3.79  && 3.10  & 134.26 && 1.24  & 142.14 && 9.27  & 424.17  && 0.69  & 448.36  && 1.34  & 38.40  \\
nrw1379 & 56638   & 16.22  & 16.88 && 3.83   & 3.86  && 2.08  & 140.10 && 2.30  & 148.15 && 15.43 & 419.56  && 1.44  & 408.74  && 0.50  & 38.32  \\
fl1400  & 20127   & 20.58  & 17.12 && 1.58   & 3.89  && 29.24 & 142.32 && 3.05  & 150.24 && 18.22 & 533.58  && 2.46  & 402.72  && 1.39  & 38.57  \\
u1432   & 152970  & 8.67   & 17.91 && 4.43   & 3.80  && 2.85  & 145.50 && 2.78  & 153.54 && 7.93  & 434.85  && 0.38  & 302.94  && 0.68  & 38.63  \\

\bottomrule
\end{tabular}
\label{Table: TSPLib1}
\end{table*}

\begin{table*}[ht]
\centering
\small
\caption{Detailed results on TSPLIB benchmark instances (Part II).}
\setlength{\tabcolsep}{0.5mm}
\begin{tabular}{c c c c c c c c c c c c c c c c c c c c c c c c}
\toprule
Instance & Optimal & \multicolumn{3}{c}{Pointerformer} & \multicolumn{3}{c}{GLOP} & \multicolumn{3}{c}{Att-GCN} & \multicolumn{3}{c}{DIFUSCO} & \multicolumn{3}{c}{LEHD RRC100} & \multicolumn{3}{c}{SIT PRC 100} & \multicolumn{3}{c}{Ours Rec 500} \\
Name     &         & Gap & Time(s) & & Gap & Time(s) & & Gap & Time(s) & & Gap & Time(s) & & Gap & Time(s) & & Gap & Time(s) & & Gap & Time(s) \\
\midrule
fl1577  & 22249   & 22.75  & 24.59 && 8.14   & 3.79  && 18.51 & 160.70 && 5.96  & 169.01 && 14.78 & 513.83  && 1.44  & 353.43  && 1.12  & 38.44  \\
d1655   & 62128   & 33.44  & 26.76 && 5.00   & 3.83  && 2.93  & 168.45 && 2.92  & 177.33 && 13.31 & 590.17  && 3.43  & 341.56  && 2.24  & 38.72  \\
vm1748  & 336556  & 22.46  & 30.80 && 5.16   & 3.85  && 2.66  & 178.41 && 2.98  & 187.55 && 10.11 & 601.56  && 0.79  & 309.68  && 0.97  & 38.67  \\
u1817   & 57201   & 36.59  & 33.79 && 6.55   & 4.00  && 4.07  & 185.52 && 3.70  & 194.75 && 9.38  & 652.42  && 1.32  & 367.50  && 1.57  & 38.58  \\
rl1889  & 316536  & 29.51  & 38.86 && 7.89   & 4.01  && 3.77  & 193.51 && 2.82  & 202.84 && 7.50  & 713.64  && 1.22  & 345.50  && 1.19  & 38.93  \\
d2103   & 80450   & 35.09  & 49.20 && 10.87  & 4.06  && 4.31  & 215.36 && 2.15  & 225.42 && 14.57 & 875.09  && 4.22  & 363.31  && 4.35  & 39.11  \\
u2152   & 64253   & 39.26  & 54.52 && 6.75   & 4.07  && 3.82  & 220.88 && 4.38  & 230.60 && 11.64 & 839.95  && 3.31  & 363.16  && 1.30  & 38.95  \\
u2319   & 234256  & 3.86   & 70.56 && 1.88   & 4.14  && 0.99  & 238.22 && 1.35  & 249.81 && 4.14  & 1032.75 && 0.22  & 320.41  && 0.25  & 39.19  \\
pr2392  & 378032  & 17.88  & 75.87 && 6.19   & 4.06  && 2.91  & 245.81 && 2.93  & 256.79 && 12.31 & 992.59  && 1.88  & 326.59  && 1.62  & 39.26  \\
pcb3038 & 137694  & 17.54  & 147.01&& 6.03   & 4.40  && 2.76  & 313.41 && 2.89  & 328.85 && 13.37 & 2313.82 && 1.59  & 364.62  && 1.35  & 39.98  \\
fl3795  & 28772   & 30.50  & 241.40&& 8.54   & 4.79  && 51.89 & 393.68 && 10.68 & 411.36 && 13.89 & 2442.83 && 2.19  & 343.84  && 2.09  & 42.16  \\
fnl4461 & 182566  & 33.99  & 308.98&& 5.07   & 5.02  && 2.81  & 466.37 && 2.84  & 484.33 && 18.95 & 3177.49 && 1.87  & 383.06  && 1.58  & 42.53  \\
rl5915  & 565530  & \multicolumn{2}{c}{OOM}   && 11.84  & 5.93  && 4.44  & 625.92 && 5.13  & 645.43 && 24.18 & 6000.23 && 3.44  & 361.73  && 2.44  & 45.30  \\
rl5934  & 556045  & \multicolumn{2}{c}{OOM}   && 10.70  & 6.04  && 4.39  & 627.18 && 4.48  & 647.45 && 24.11 & 5755.70 && 3.93  & 376.41  && 2.41  & 46.58  \\
rl11849 & 923288  & \multicolumn{2}{c}{OOM}     && 10.04  & 13.09 && \multicolumn{2}{c}{OOM}      && 4.86  & 1321.03&& 38.03 & 32915.73&& 4.85  & 362.65  && 2.48  & 68.18  \\
usa13509& 19982859& \multicolumn{2}{c}{OOM}   && 6.08   & 15.82 && \multicolumn{2}{c}{OOM}      && 10.68 & 1511.73&& \multicolumn{2}{c}{OOM}       && 8.29  & 414.80  && 3.40  & 118.16 \\
brd14051& 469385  & \multicolumn{2}{c}{OOM}   && 5.46   & 16.78 && \multicolumn{2}{c}{OOM}      && 4.26  & 1568.55&& \multicolumn{2}{c}{OOM}       && 4.04  & 429.64  && 2.88  & 107.49 \\
d15112  & 1573084 & \multicolumn{2}{c}{OOM}   && 5.39   & 19.80 && \multicolumn{2}{c}{OOM}      && 3.65  & 1693.48&& \multicolumn{2}{c}{OOM}       && 3.81  & 440.61  && 2.66  & 105.44 \\
d18512  & 645238  & \multicolumn{2}{c}{OOM}   && 5.57   & 27.52 && \multicolumn{2}{c}{OOM}      && 3.43  & 2116.37&& \multicolumn{2}{c}{OOM}       && 4.03  & 400.12  && 2.93  & 148.68 \\

\bottomrule
\end{tabular}
\label{Table: TSPLib2}
\end{table*}

\end{document}